\documentclass{article}

\usepackage[preprint]{neurips_2026}

\usepackage[utf8]{inputenc} % allow utf-8 input
\usepackage[T1]{fontenc}    % use 8-bit T1 fonts
\usepackage[breaklinks=true]{hyperref}       % hyperlinks
\let\oldcitet\citet
\let\oldcitep\citep
\renewcommand{\citet}[1]{\mbox{\oldcitet{#1}}}
\renewcommand{\citep}[1]{\mbox{\oldcitep{#1}}}
\usepackage{url}            % simple URL typesetting
\usepackage{booktabs}       % professional-quality tables
\usepackage{amsfonts}       % blackboard math symbols
\usepackage{nicefrac}       % compact symbols for 1/2, etc.
\usepackage{microtype}      % microtypography
\usepackage{xcolor}         % colors
\usepackage{tabularx}
\usepackage{amsmath}
\usepackage{graphicx}
\usepackage{subcaption}
\hypersetup{hidelinks}
\usepackage{multirow}
\usepackage{caption}
\usepackage[section]{placeins}  % \FloatBarrier per (sub)section keeps figures with their headings

\title{Sparsity Moves Computation: How FFN Architecture Reshapes Attention in Small Transformers
}

\author{
  Gabriel Smithline \\
  University of Michigan \\
  \texttt{gsmithl@umich.edu}
  \And
  Chris Mascioli \\
  University of Michigan \\
  \texttt{cmasciol@umich.edu}
}

\begin{document}

\maketitle

\begin{abstract}
Architectural choices inside the Transformer feedforward network (FFN) block do not merely affect the block itself; they reshape the computations learned by the rest of the model. We study this effect in one-layer Transformers trained on digit addition with carry, modular arithmetic, and histogram counting. Comparing dense FFNs, gated linear units (GLUs), mixture-of-experts (MoE), and MoE-GLUs, we find that sparse MoE routing can shift computation from FFN to attention, with the strongest ablation-visible effect on carry-based addition. We decompose this redistribution into reduced per-token FFN capacity and sparse partitioning across experts. Critically, frozen random routing nearly matches learned routing, suggesting that redistribution is driven largely by architectural sparsity rather than router-learned specialization. As a secondary finding, GLU-style multiplicative gating rotates task-relevant Fourier structure out of the per-neuron basis and into distributed subspaces, making neuron-level interpretability less informative while preserving structured computation. We validate these conclusions with random-routing, narrow-FFN, and top-2 MoE controls, plus parameter-matching, activation-function, and width-scaling analyses. Together, these results show that local FFN design choices can have nonlocal consequences for Transformer computation.
\end{abstract}

\vspace{-0.7em}

\begin{center}
  \includegraphics[width=0.74\linewidth]{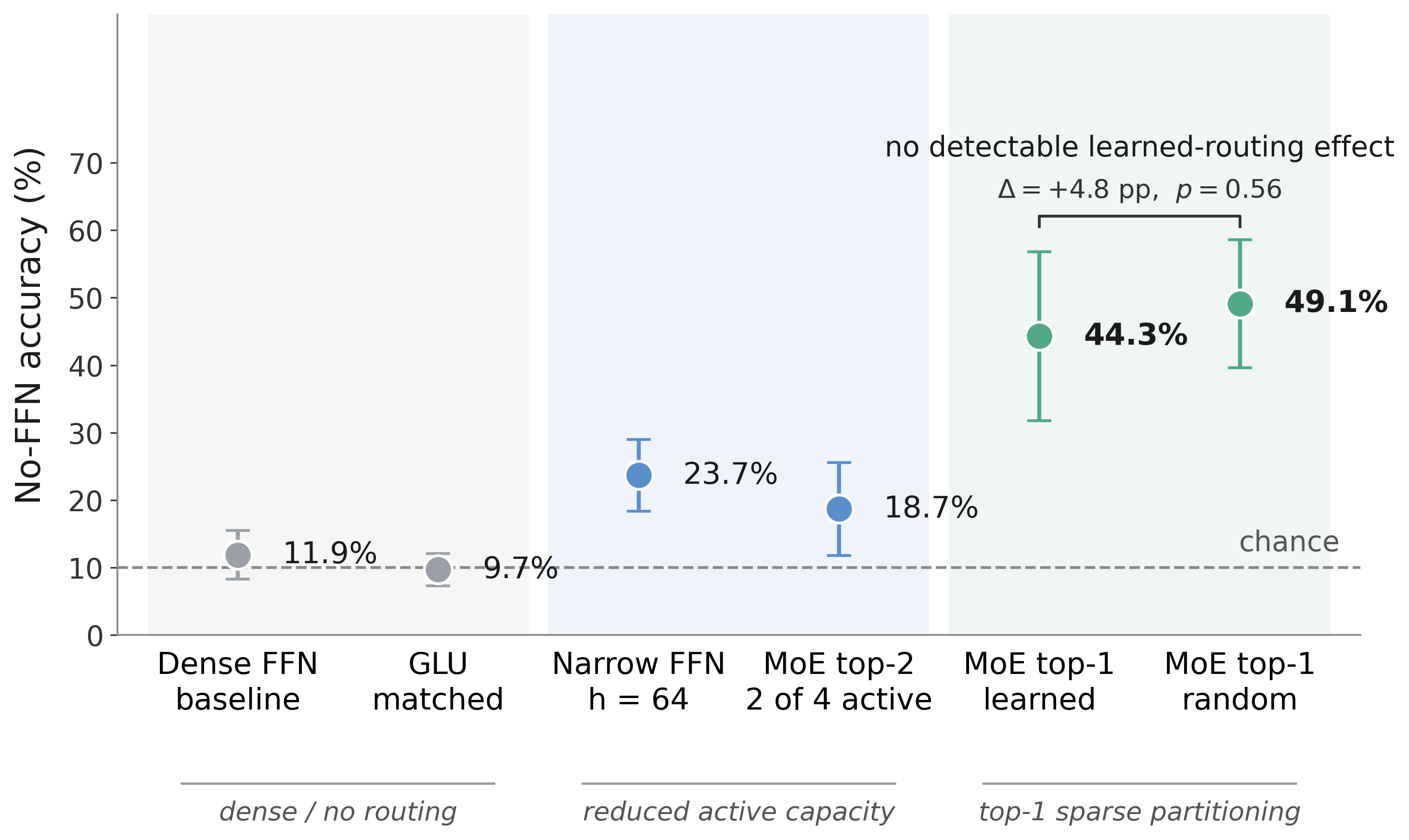}
  \captionof{figure}{\textbf{Sparse partitioning, not learned routing, drives FFN-to-attention redistribution.}
  Mean no-FFN accuracy on add-7 over 5 seeds. Narrow FFN isolates reduced per-token capacity; top-2 relaxes the bottleneck; random routing preserves sparse partitioning without learned router adaptation and retains comparable no-FFN accuracy to learned routing. Full decomposition in Sec.~\ref{sec:routing-causal}.}
  \label{fig:headline}
\end{center}

\vspace{-0.7em}

\section{Introduction}

Architectural choices inside the Transformer feedforward network (FFN) block are often treated as interchangeable local modifications. Dense FFNs, gated linear units (GLUs), and mixture-of-experts (MoE) layers occupy roughly the same position in the residual stream, yet they perform different computations and impose different computational constraints on each token.
In particular, MoE layers keep total FFN capacity high while reducing the active FFN capacity available to any single token. This suggests a nonlocal possibility: changing the FFN architecture may not only change the FFN computation, but also change what attention learns to compute.  We study this possibility in the controlled setting of one-layer Transformers trained on algorithmic tasks. Mechanistic interpretability has shown that such models learn structured circuits, including Fourier-based algorithms for modular addition \citep{nanda2023progress, Zhong2023TheCA}, position-specific routines for multi-digit addition \citep{quirke2023understanding}, and architecture-dependent divisions of labor between attention and FFN in counting tasks \citep{Behrens2024CountingIS}. We train four FFN variants: dense FFN, GLU, MoE, and MoE-GLU, on three tasks: digit-by-digit addition of 7 with carry propagation, modular addition over $\mathbb{Z}_{113}$, and histogram counting. We then trace how changing the FFN changes what attention learns. 

Our main findings are as follows:
\begin{itemize}
\item \textbf{MoE shifts computation from FFN to attention.} With FFN ablated, MoE retains $44\%$ accuracy on add-7 vs.\ $12\%$ for parameter-matched dense FFN and $9\%$ for GLU. The effect is strongest when the task is within attention's expressive capacity (Sec.~\ref{sec:redistribution}).

\item \textbf{Learned routing is not necessary for redistribution.}
A random-routing control, with router weights frozen at initialization, produces nearly the same shift toward attention as learned routing. While prior work has shown that random expert assignment can be competitive with learned routing \citep{zuo2021taming}, we use random routing mechanistically: to isolate sparse partitioning from router-learned specialization. The FFN-to-MoE gap decomposes into reduced per-token FFN capacity and sparse partitioning across experts, with no detectable additional contribution from learned router adaptation (Sec.~\ref{sec:routing-causal}).

\item \textbf{GLU rotates structure out of the per-neuron basis.} GLU does not produce the MoE redistribution effect, but it changes the format of the learned computation. Per-neuron Fourier concentration drops, while PCA recovers Fourier-aligned structure in a low-dimensional subspace. Thus task information is preserved but becomes less visible to neuron-level interpretability tools (Sec.~\ref{sec:glu}).
\end{itemize}

\section{Background and Notation}

\paragraph{Architecture.} We study a single-layer transformer: an input sequence is mapped to embeddings $e_i = E x_i + p_i$ (learned positional encodings), then passed through multi-head self-attention and a feed-forward network with residual connections around each block. For each head $h$, the attention output is $\text{Attn}_h(X) = \text{softmax}(Q_h K_h^\top / \sqrt{d_k})\, V_h$ with $Q_h, K_h, V_h$ linear projections of $X$. The multi-head output is $\text{MHA}(X) = [\text{Attn}_1; \ldots; \text{Attn}_H] W_O$, and the residual stream evolves as $X' = X + \text{MHA}(X)$ followed by $X'' = X' + \text{FFN}(X')$. Output logits are produced by projecting through $W_U = E^\top$ (tied embeddings \citep{press2017using, inan2017tying}) for add-7; for modular addition and histogram the output head is untied because the output classes differ from the input vocabulary or use a separate cyclic-group target.

\paragraph{Model variants.} We compare four variants:
\begin{itemize}
\item \textbf{FFN}: $\text{FFN}(x) = W_{\text{down}}\, \sigma(W_{\text{up}}\, x)$.
\item \textbf{GLU} \citep{dauphin2017language, shazeer2020glu}: $\text{GLU}(x) = W_{\text{down}}(\sigma(W_{\text{gate}}\, x) \odot W_{\text{up}}\, x)$, introducing a bilinear interaction via the multiplicative gate.  We set $h_{\text{glu}}{=}\tfrac{2}{3}h_{\text{dense}}$ to total parameter match the dense baseline (since GLU has three weight matrices vs. dense's two).
\item \textbf{MoE} \citep{shazeer2017outrageously}: a router $p(x) = \text{softmax}(W_r x)$ selects the top-1 of $E$ expert FFNs, with an auxiliary load-balancing loss \citep{fedus2022switch}.  Each expert has $1/E$ the hidden dimension, and total FFN parameters match the dense baseline. 
\item \textbf{MoE-GLU}: the same routing with GLU experts at $h_E{=}\tfrac{2}{3}h_{\text{dense}}/E$ for the results. This is total parameter matched as GLU (total expert width sums to $\tfrac{2}{3}h_{\text{dense}}$, matching dense FFN total params within 1-2\%). Per-active capacity is $1/E$ of dense FFN.

\end{itemize}
Top-2 controls (App.~\ref{app:topk}) are applied to both MoE variants, and use the same architecture with two active experts per token. This isolates the routing-$k$ axis at fixed architecture. We test both SiLU \citep{elfwing2017sigmoidweighted,ramachandran2017searching} and GELU \citep{Hendrycks2016GaussianEL} as the activation $\sigma$ across all GLU and MoE-GLU experiments; the choice does not materially affect any of our findings (App.~\ref{app:silu}, App.~\ref{app:activation-robustness}).

\paragraph{Mechanistic interpretability of algorithmic tasks.} Our work builds on a body of mechanistic analyses of algorithmic tasks. \citet{nanda2023progress} showed that transformers trained on modular addition learn discrete Fourier circuits, with individual neurons responding to specific frequencies. \citet{quirke2023understanding} found that multi-digit addition decomposes into parallel per-digit streams with position-specific algorithms. Our add-7 task extends their setup to ask how FFN architecture changes these circuits.

\paragraph{The attention-FFN division of labor.} \citet{Behrens2024CountingIS} identified two counting strategies in small transformers, \emph{relation-based} (attention-driven) and \emph{inventory-based} (FFN-driven), showing that minor architectural changes shift which strategy is adopted. This finding motivates our central question.   Parallel work by \citet{dong2025random} showed that transformers form specialized circuits even with frozen random attention, establishing that attention and FFN play distinct but substitutable roles. Our frozen-component experiments (App.~\ref{app:frozen}) test whether each component can support the task when the other is held fixed at random initialization. These controls complement the main architectural comparisons, which vary the FFN architecture and show that the learned attention and FFN division of labor shifts as the FFN bottleneck changes.

\paragraph{Mechanistic analysis of MoE.}
Mechanistic studies of MoE remain comparatively sparse relative to dense FFNs. 
Recent work analyzes pretrained MoE LLMs by attributing factual knowledge to routed experts \citep{li2025decoding}, decomposing cross-layer contributions to router decisions \citep{liunderstanding}, and tracing how MoE and dense models acquire knowledge over pretraining \citep{wang2026deconstructing}. 
Together, these studies suggest that MoE computation involves expert collaboration, cross-layer routing dependencies, and stable sparse knowledge organization.  Our work is complementary: rather than explaining knowledge attribution or routing in large pretrained MoEs, we examine controlled one-layer transformers on algorithmic tasks to isolate how MoE architecture changes the learned division of labor between FFN and attention. 
Our claim is not that MoEs universally rely less on FFNs, but that sparse expert partitioning and reduced per-token FFN capacity can cause a redistribution of algorithmic computation into attention. Our narrow-FFN and random-routing controls show that this redistribution can arise from reduced per-token FFN capacity and sparse partitioning even without learned expert specialization. 

% \paragraph{Expressiveness of attention.} \citet{Weiss2021ThinkingLT} formalized what attention can compute via RASP. \citet{Hahn2019TheoreticalLO} and \citet{Merrill2021SaturatedTA} showed that self-attention can express disjunctive but not conjunctive composition. We use these results to explain why redistribution works on add-7 (per-digit addition is within attention's reach) but not modular addition (Fourier computation exceeds it).

\paragraph{Random routing and attention--FFN division of labor.}
Prior work has shown that stochastic or fixed random expert activation can be effective as an optimization, regularization, or scaling mechanism \citep{zuo2021taming, chen2023sparse}. We instead use random routing as a mechanistic intervention: freezing the router tests whether learned expert specialization is necessary for the FFN-to-attention shift. Random routing largely preserves the shift, suggesting that sparse expert partitioning is sufficient in our setting. 
This complements work showing that small Transformers solve counting-like tasks through a delicate attention--FFN division of labor \citep{Behrens2024CountingIS}.  Our work shows that MoE-style FFN sparsity can systematically reshape that division of labor.

\paragraph{GLUs and privileged bases.}
Our GLU results relate to work on whether MLP representations have a privileged basis. 
Superposition analyses argue that standard nonlinearities can make individual neurons meaningful units of analysis \citep{elhage2022toy}, while work on bilinear MLPs shows that GLU-like architectures may expose structure more naturally through tensor or spectral decompositions of weights than through individual neurons \citep{pearce2024bilinear}. 
Our finding is reciprocal: GLU-style multiplicative gating preserves task-relevant structure in controlled algorithmic tasks, but rotates it out of the per-neuron basis. This causes neuron-level probes to understate the learned computation.

\section{Experimental Setup}
\label{sec:experimental-setup}

\begin{figure}[htbp]
  \centering
  \includegraphics[width=0.89\linewidth]{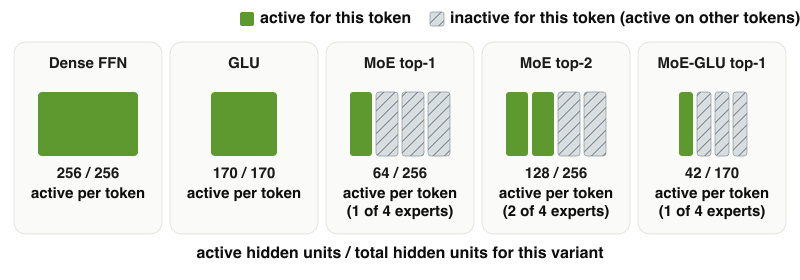}
  \caption{\textbf{Active FFN capacity per token across architecture variants.}
  Green blocks are active for a token; hatched gray blocks are inactive for that token but active for other routed tokens. 
  MoE top-$k$ activates only routed experts, producing the per-token FFN bottleneck studied in Sec.~\ref{sec:redistribution}. 
  Widths are total-parameter matched; full details are in Tab.~\ref{tab:ffn-params}.}
  \label{fig:capacity}
\end{figure}

\label{sec:setup}

We study three algorithmic tasks chosen to stress different balances of attention to FFN computation. All use a one-layer transformer with $d_{\text{model}}{=}64$ (add-7) or $d_{\text{model}}{=}128$ (modular addition, histogram), 4 attention heads, and four FFN variants: FFN, GLU, MoE ($E{=}4$ experts, top-1 routing in main results), and MoE-GLU. We additionally run all MoE and MoE-GLU experiments at top-2 routing as an ablation (App.~\ref{app:topk}). We train without RMSNorm for primary analyses  (App.~\ref{app:norm}). All four variants are total parameter matched to FFN within $1-2\%$. Each MoE expert has $1/E$ the hidden dimension, GLU uses $\tfrac{2}{3}$ the hidden dimension to absorb its third weight matrix ($3 d \cdot \tfrac{2}{3} h = 2 d h$). All variants are parameter-matched by construction; full details and audits are in App.~\ref{app:architecture-details}.

%Concretely: $h_{\text{glu}}{=}170$ for $d{=}64$ ($32{,}640$ vs.\ FFN's $33{,}088$ params) and $h_{\text{glu}}{=}340$ for $d{=}128$ ($130{,}560$ vs.\ $131{,}712$). MoE-GLU is total-parameter-matched-as-GLU: total expert width is $\sum_e h_E = \tfrac{2}{3} h_{\text{dense}}$, with each expert at $h_E = \tfrac{2}{3} h_{\text{dense}} / E$ ($h_E{=}42$ on add-7, $h_E{=}85$ on modadd and histogram). Total MoE-GLU FFN parameters match FFN within 1-2\%, and per-token (active) capacity is $1/E$ of dense.  The redistribution effect we report is therefore measured under total-parameter matching. MoE variants have no more total FFN parameters than the dense baseline, while each token accesses strictly less active FFN parameters (Tab.~\ref{tab:ffn-params}). Top-2 controls (App.~\ref{app:topk}) train an architecturally identical model with $k{=}2$ routing, isolating the routing-$k$ axis at fixed architecture. All experiments use 5 seeds. Full task descriptions, architecture details, and analysis method definitions are in App.~\ref{app:setup-details}.

% We total-parameter match plain MoE to dense FFN rather than FLOP-matching as in the scaling-MoE literature see App.~\ref{app:param-vs-flop} for the full rationale.

\paragraph{Add-7.} Compute $x + 7$ digit by digit on 3-digit numbers in reversed order (ones first), so that carries propagate left to right through the sequence. Each output position requires one of three operations: add $7$ at the ones position, add $1$ when a carry propagates through a prefix of 9s, or pass the digit through unchanged. The task therefore tests positionally local digit transformations gated by a non-local carry trigger, following \citet{quirke2023understanding}. We sample uniformly from all $1000$ three-digit inputs each batch. Evaluation uses the same uniform distribution, and Sec.~\ref{sec:redistribution} stratifies accuracies by carry-chain length.

\paragraph{Modular addition.} Solve $(a + b) \bmod 113$ from input $[a, b, \mathord{=}]$, following \citet{nanda2023progress}. The fixed $113 \times 113$ input grid has no positional or compositional structure to decompose, so the task tests whether the model can learn the cyclic group $\mathbb{Z}_{113}$ as a global algebraic relation.  This requires nonlinear Fourier circuits that attention alone cannot compute.

\paragraph{Histogram counting.} Given $L{=}10$ tokens from alphabet size $T{=}32$, compute each token's frequency in the sequence, following \citet{Behrens2024CountingIS}. Histogram counting requires both sequence-level selection and token-specific readout, namely the model must identify positions containing the same token and map that information to a count. Prior work shows that small Transformers can solve such tasks using relation-style or inventory-style strategies, depending on architectural details \citep{Behrens2024CountingIS}. We use histogram to distinguish \emph{internal} shifts in component roles from \emph{ablation-visible} redistribution, where attention absorbs enough FFN work for the model to remain functional when the FFN is removed. In our experiments, histogram shows little ablation-visible redistribution because attention and the FFN remain complementary rather than substitutable.

\paragraph{Analysis methods.} Our primary redistribution measure is component ablation: we zero either the attention output or the FFN output and measure per-token accuracy. This captures \emph{ablation-visible} redistribution: cases where one component has absorbed enough computation that the model remains partially functional when the other is removed. We support this with direct logit attribution, activation patching \citep{meng2022locating}, linear probes, Fourier analysis, and routing analysis, which can reveal internal strategy shifts even when ablation accuracy remains low. Details are in App.~\ref{app:analysis-details}.

\paragraph{Evaluation metrics.}
At each output position $t$, we take the $\arg\max$ of the logits and average $\mathbf{1}[\hat y_t = y_t]$ over output positions and examples, reporting \emph{per-token output accuracy}. The output region is task-specific: for add-7, we average over the four output digits $o_0, o_1, o_2, o_3$ (Ones, Tens, Hundreds, Overflow), excluding the trailing EOS / PAD positions which are trivially predicted. For modular addition, we evaluate the single output position following the $\mathord{=}$ token, and for histogram counting, we average over all $L=10$ output positions.

%%%%%%%%%%%%%%%%%%%%%%%%%%%%%%%%%%%%%%%%%%%%%%%%%%%%%%%%%%%%
\section{Results}
%%%%%%%%%%%%%%%%%%%%%%%%%%%%%%%%%%%%%%%%%%%%%%%%%%%%%%%%%%%%

\subsection{MoE Redistributes Computation from FFN to Attention}
\label{sec:redistribution}

We begin with our central finding: replacing FFN with MoE changes where computation happens.

\begin{figure}[htbp]
  \centering
  \begin{subfigure}[t]{0.49\linewidth}
    \centering
    \includegraphics[width=\linewidth]{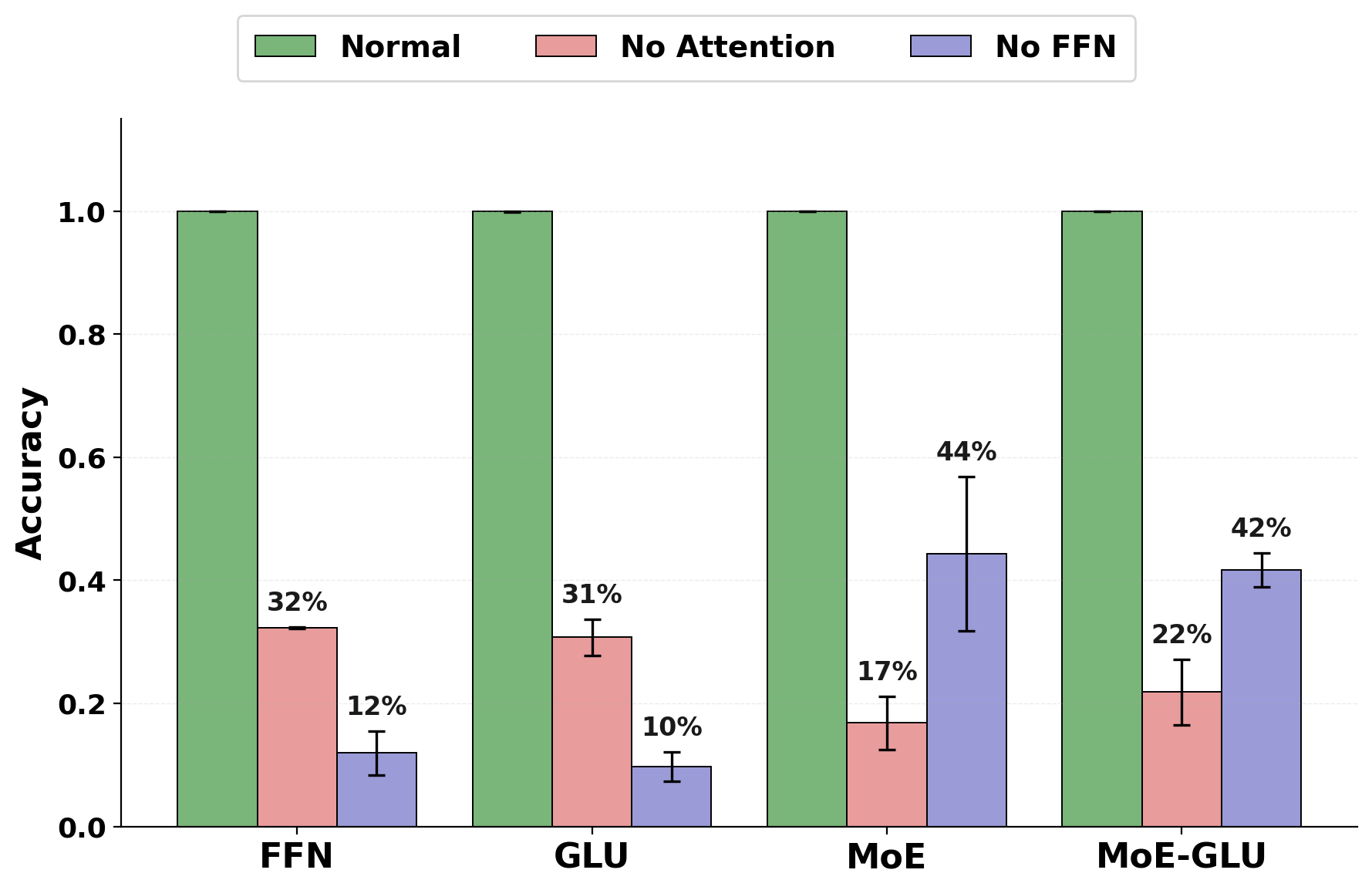}
    \caption{Add-7 (3-digit). MoE retains $44.3\%$ without FFN vs.\ $11.9\%$ for FFN at matched widths.}
    \label{fig:ablation-add7}
  \end{subfigure}
  \hfill
  \begin{subfigure}[t]{0.49\linewidth}
    \centering
    \includegraphics[width=\linewidth]{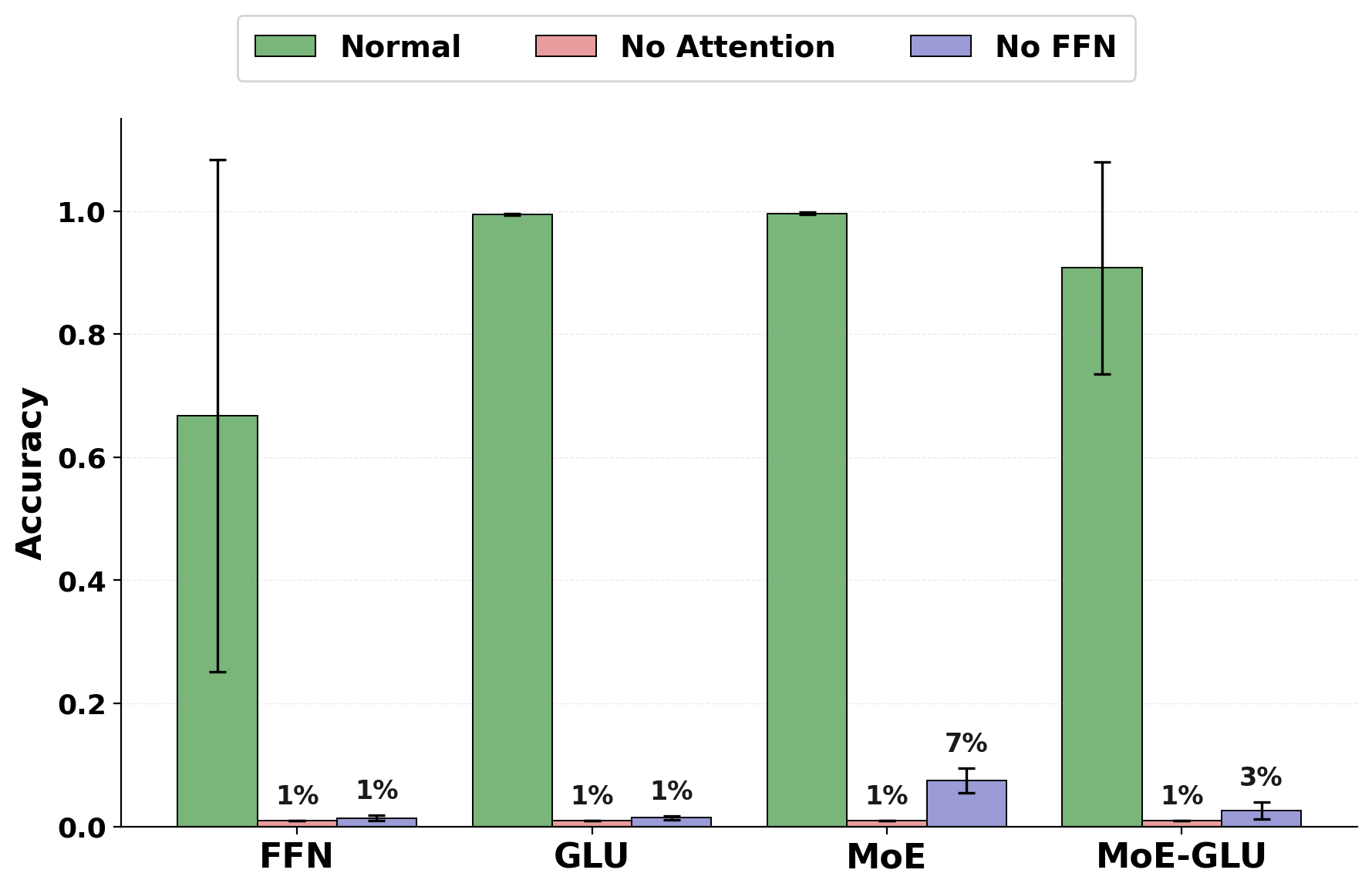}
    \caption{Modular addition ($\mathbb{Z}_{113}$). MoE retains $7.5\%$ vs.\ $1.3\%$ for dense at matched widths.}
    \label{fig:ablation-modadd}
  \end{subfigure}
  \caption{\textbf{Component ablation across tasks at matched parameter widths} All variants are total-parameter-matched within 1-2\%: FFN, GLU with $h{=}\tfrac{2}{3}h_{\text{dense}}$, MoE with $h_E{=}h_{\text{dense}}/E$, and MoE-GLU with $h_E{=}\tfrac{2}{3}h_{\text{dense}}/E$, so total expert width is $\tfrac{2}{3}h_{\text{dense}}$ (Tab.~\ref{tab:ffn-params}). At parity, MoE retains much higher no-FFN accuracy on tasks with ablation-visible redistribution pressure: largest on add-7 ($+32.4$\,pp over FFN), smaller but consistent on modular addition ($+6.2$\,pp), and not ablation-visible on histogram ($+0.5$\,pp; App.~\ref{app:histogram-errors} shows internal strategy shifts).  MoE-GLU follows the same trend (add-7 $+29.8$\,pp). Accuracy is the fraction of the four output digits predicted correctly.}
  \label{fig:component-ablation}
\end{figure}

\paragraph{The redistribution effect.}
With FFN zeroed, MoE retains $44.3 \pm 12.5\%$ accuracy on add-7 vs.\ $11.9 \pm 3.6\%$ for FFN at matched widths (Fig.~\ref{fig:ablation-add7}). GLU also retains only $9.7 \pm 2.4\%$, while MoE-GLU retains $41.7 \pm 2.8\%$: the dividing line is sparse routing, not the choice of FFN nonlinearity.

The effect is not explained by extra total parameters. At per-active (FLOP) parity with $4\times$ dense total parameters, MoE-GLU retains $18.3 \pm 5.6\%$ no-FFN accuracy on add-7 (App.~G.5). While still above FFN's $11.9\%$, per-active matching preserves only a small fraction of the total-parameter-matched no-FFN gap ($6.4$\,pp vs.\ $29.8$\,pp). This indicates that active expert width and sparse partitioning interact, rather than contributing strictly additively across the MoE-GLU controls. 

Nor is the effect a strict capacity limitation. Frozen-component training (App.~\ref{app:frozen}) shows that MoE's FFN alone reaches $100\%$ on add-7, confirming that it can solve the task independently. Instead, MoE's routing reduces per-token FFN capacity to ${\sim}1/E$, and attention compensates by absorbing computation it would otherwise leave to the FFN. Direct logit attribution supports this interpretation: before ablation, attention contributes ${\sim}60\%$ of correct logits in MoE vs.\ ${\sim}40\%$ in FFN (App.~\ref{app:dla}).

% \paragraph{What does the redistribution depend on?}
% Three controls (top-2 routing, a narrow dense FFN at $1/E$ width, and random non-learned routing) jointly decompose the effect into a per-token capacity component and a sparse-partitioning component, and show that the partitioning does not have to be learned. We defer the full decomposition to Sec.~\ref{sec:routing-causal}.

\begin{figure}[htbp]
  \centering
  \includegraphics[width=0.85\linewidth]{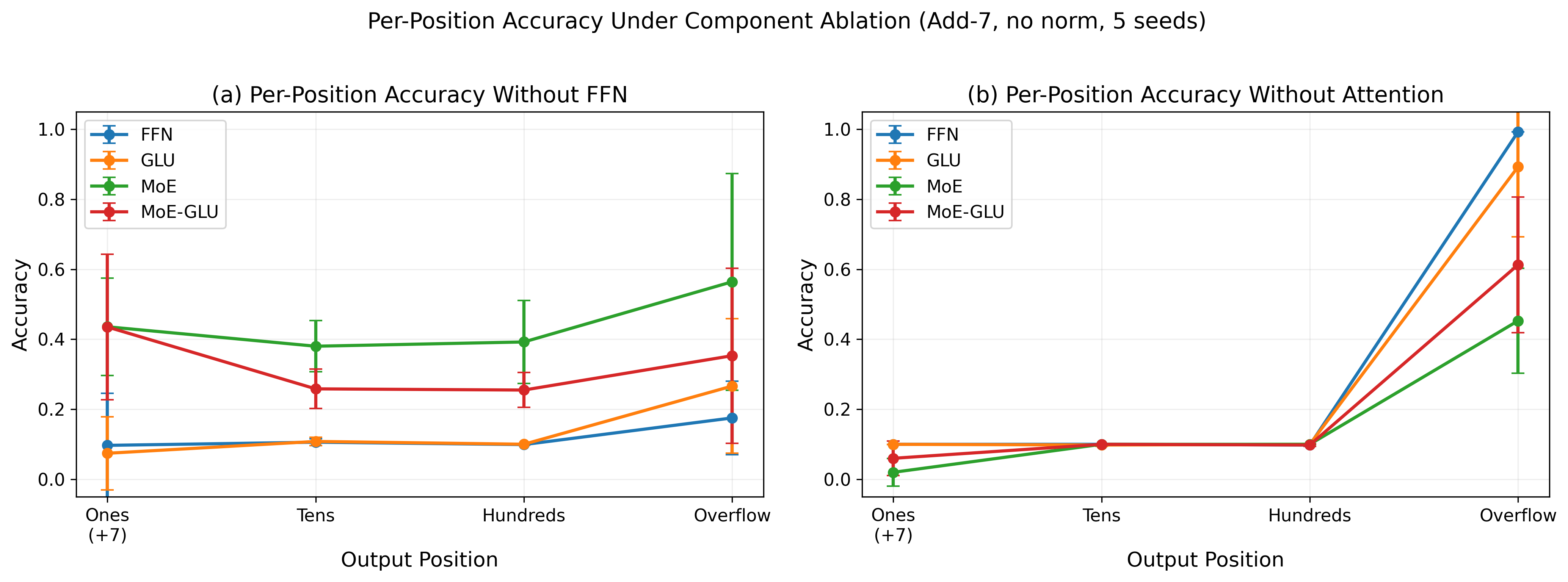}
  \caption{\textbf{Position-dependent redistribution on add-7} (5 seeds). \textbf{Left}: without FFN, MoE's attention handles all positions at 38-56\%, while FFN collapses to ${\sim}10\%$ (chance). \textbf{Right}: without attention, FFN retains high overflow accuracy while MoE's FFN drops substantially. Because overflow is highly imbalanced, this no-attention result should be interpreted as localization evidence rather than proof that the FFN computes overflow aggregation.}
  \label{fig:per-position}
\end{figure}

% \paragraph{Position-dependent redistribution.}
% The effect is broadly distributed across output positions (Fig.~\ref{fig:per-position}). Without FFN, MoE's attention handles all positions at 38-56\% accuracy, while FFN's attention collapses to ${\sim}10\%$. The advantage is largest at the overflow position, where carry information must be aggregated across the full input. Conversely, without attention, FFN retains 99\% at overflow while MoE's FFN drops to 45\%, having offloaded that work. This pattern holds across all carry-chain lengths (App.~\ref{app:ablation-summary}).

\paragraph{Position-dependent redistribution.}
The effect is broadly distributed across output positions (Fig.~\ref{fig:per-position}). Without FFN, MoE's attention handles all positions at 38-56\% accuracy, while FFN's attention collapses to ${\sim}10\%$. The advantage is largest at the overflow-sensitive position, where correct behavior depends on carry information from the full input. Conversely, without attention, FFN retains high overflow accuracy while MoE's FFN drops substantially. Because overflow events are highly imbalanced, high no-attention overflow accuracy can partly reflect the majority-class baseline rather than learned carry aggregation. We therefore interpret the position-wise result as localization evidence: ablation-visible redistribution is concentrated where non-local carry information is most relevant. This pattern holds across all carry-chain lengths (App.~\ref{app:ablation-summary}).

\paragraph{Cross-task validation: modular addition.}
On modular addition, no-FFN accuracy is low for all variants (MoE: $7.5\%$, FFN: $1.3\%$), but MoE still retains ${\sim}5{\times}$ more (Fig.~\ref{fig:ablation-modadd}). Modular addition requires nonlinearities, so the residual MoE advantage is small in absolute terms but consistent in direction with add-7.

% \paragraph{Cross-task validation: histogram counting.}
% On histogram counting, redistribution is essentially absent (${<}1$\,pp gap between any pair of variants, all four variants land within $11.5$-$12.0\%$ for FFN/MoE and $9.8$-$10.5\%$ for GLU/MoE-GLU at matched widths, App.~\ref{app:histogram-errors}). Histogram is roughly solvable by either component alone (App.~\ref{app:frozen}: $98$-$99\%$ with frozen attention, $94$-$97\%$ with frozen FFN), so neither component faces a bottleneck the other must compensate for. But what is different is histogram's low redistribution is therefore imposed by the task, not by the FFN architecture, a negative control confirming that the redistribution we observe on add-7 and modadd is architecture-dependent rather than a measurement artifact. 

\paragraph{Cross-task validation: histogram counting.}
On histogram counting, MoE does not produce comparable \emph{ablation-visible} redistribution. No-FFN accuracy remains near chance for all variants (App.~\ref{app:histogram-errors}). This does not mean the FFN is unused. Rather, histogram separates internal shifts in component roles from functional robustness under ablation. Following \citet{Behrens2024CountingIS}, counting can decompose into selection-like work, naturally supported by attention, and token-inventory or nonlinear readout work, which is naturally supported by the FFN. Our appendix diagnostics show that, in the FFN-family, MoE increases the attention-side count signal. Attention lift rises from $+0.572$ for dense FFN to $+0.722$ for MoE (App.~\ref{app:histogram-errors}). However, the FFN remains necessary, so this internal shift does not translate into no-FFN robustness. Thus histogram is a non-substitutable control: sparse partitioning can change internal strategy without making attention able to replace the FFN.

% Error analysis (App.~\ref{app:histogram-errors}) reveals that without FFN, all variants collapse on extreme counts (1, 2, 8, 9: ${<}5\%$ accuracy) while partially preserving mid-range counts (4-5: up to $64\%$ for MoE), consistent with softmax normalizing away count magnitude.

%-------------------------------------------------------------------
\subsection{Sparse Partitioning, Not Learned Routing, Drives Redistribution}
\label{sec:routing-causal}

Sec.~\ref{sec:redistribution} established that MoE shifts substantial work from FFN to attention, producing a $+32.4$\,pp gap on add-7 between matched FFN ($11.9\%$ no-FFN) and MoE ($44.3\%$). We now decompose this gap to isolate which property of MoE produces it. Three candidates are available a priori:

\begin{itemize}
\item \textbf{(i) Per-token capacity reduction.} Each expert sees only $1/E$ of the hidden dimension, so any single token has access to less FFN capacity than under plain FFN.
\item \textbf{(ii) Sparse partitioning across experts.} Different inputs are routed to different experts, so the FFN implements different effective functions across the input distribution.
\item \textbf{(iii) Learned routing.} The router $W_r$ adapts during training, so the partitioning is task-aware rather than arbitrary.
\end{itemize}

Three controls separate these in order. A \emph{narrow FFN} isolates (i) from (ii) by reducing per-token capacity without introducing partitioning. \emph{Random routing} isolates (ii) from (iii) by preserving partitioning while freezing it at initialization. \emph{Top-2 routing} partially relaxes the per-token bottleneck at fixed total architecture by doubling the active expert width, providing a final consistency check. Together, these controls operationally decompose the $+32.4$\,pp FFN-to-MoE redistribution gap (Tab.~\ref{tab:random-routing}) into a per-token-capacity component and a sparse-partitioning component, while showing that learned routing contributes no detectable additional component.

A per-head decomposition shows that the same causal ladder appears within attention itself: at carry-propagation positions, narrow FFN and MoE variants increasingly concentrate attention-side correct-logit contribution in the top-ranked head, with random-routing MoE matching learned MoE in both Rank-1 share and top-head magnitude (App.~\ref{app:per-head-dla}).

\paragraph{Narrow FFN: capacity reduction accounts for $+12$\,pp.}
An FFN with hidden dim equal to one MoE expert's width ($h{=}64$, $1/E$ of the matched FFN) isolates candidate (i) from (ii): per-token capacity is reduced, but no partitioning is introduced. The narrow FFN retains $23.7 \pm 5.3\%$ no-FFN accuracy on add-7, compared to $11.9 \pm 3.6\%$ for matched FFN and $44.3 \pm 12.5\%$ for MoE (learned routing) at the same per-active capacity (Tab.~\ref{tab:random-routing}). Reduced per-token capacity alone accounts for $+12$\,pp of the gap, leaving the $+21$\,pp attributable to sparse partitioning across multiple experts.

% \paragraph{Random routing: the remaining $+25$\,pp does not require learning.}
% We now isolate candidate (ii) from candidate (iii). We train MoE variants with router weights frozen at initialization: $W_r$ is sampled once and never updated, so each input is deterministically assigned to an expert by a fixed random projection of its embedding. Different inputs are still routed to different experts (the partitioning is preserved), but the routing function does not adapt. Experts and the rest of the network train normally.

\begin{table}[h]
\centering
\small
\begin{tabular}{lccc}
\toprule
\textbf{Variant (add-7)} & Normal & No-attn & No-FFN \\
\midrule
 FFN                        & $100.0\%$ & $32.3 \pm 0.1\%$ & $11.9 \pm 3.6\%$ \\
Narrow FFN ($h{=}64$, no routing) & $100.0\%$ & $19.8 \pm 5.7\%$ & $23.7 \pm 5.3\%$ \\
MoE (learned routing)            & $100.0\%$ & $16.8 \pm 4.3\%$ & $44.3 \pm 12.5\%$ \\
\textbf{MoE (random routing)}    & $100.0\%$ & $15.8 \pm 5.1\%$ & $\mathbf{49.1 \pm 9.5\%}$ \\
MoE-GLU (learned routing)        & $100.0\%$ & $21.8 \pm 4.9\%$ & $32.5 \pm 6.0\%$ \\
\textbf{MoE-GLU (random routing)}& $100.0\%$ & $25.3 \pm 4.0\%$ & $\mathbf{36.0 \pm 7.2\%}$ \\
\bottomrule
\end{tabular}
\caption{\textbf{Mechanism-isolation controls on add-7} (5 seeds each). 
The narrow FFN matches a single MoE expert's per-token width without sparse partitioning, while random routing preserves sparse partitioning but removes learned router adaptation. 
Random routing preserves no-FFN accuracy for both MoE and MoE-GLU, supporting the interpretation that the redistribution effect comes from sparse partitioning rather than learned routing.  Values use the digit-only metric averaged over $o_0$ to $o_3$.}

\label{tab:random-routing}
\end{table}

For plain MoE, no-FFN accuracy is $49.1 \pm 9.5\%$ under random routing vs.\ $44.3 \pm 12.5\%$ under learned routing (Welch $p{\approx}0.56$); for MoE-GLU at default width, $36.0 \pm 7.2\%$ vs.\ $32.5 \pm 6.0\%$ ($p{\approx}0.48$). Although the router itself is frozen, the rest of the network can still adapt around the induced partition. This adaptation is sufficient to match learned routing on the redistribution measure: candidate (iii) contributes no detectable redistribution beyond what sparse partitioning alone produces. We report expert-operation NMI analyses in App.~\ref{app:specialization}.

\paragraph{Top-2 routing: relaxing the bottleneck eliminates the effect.}
As a final consistency check, top-2 uses the same architecture as top-1 with $k{=}2$ routing, doubling per-active capacity at fixed architecture. At fixed $h_E$, no-FFN accuracy drops from $44.3\%$ to $18.7\%$ for MoE and from $32.5\%$ to $12.8\%$ for MoE-GLU, the latter approaching matched GLU's $\approx 9\%$ level (App.~\ref{app:topk}). The redistribution effect is tied to the bottleneck: relax the bottleneck and the effect goes away, confirming the decomposition above.

\paragraph{Modular addition: routing affects grokking, not redistribution.}
On modular addition, learned and random routing remain indistinguishable on the no-FFN redistribution gap ($1$--$8\%$ across variants), even though learned routing affects optimization dynamics. 
For plain MoE, learned routing groks faster and more reliably than random routing: learned MoE groks in all seeds, while random routing groks in $4/5$ (one seed fails to grok in 40k epochs) seeds and takes longer when it does (${\sim}33$k vs.\ ${\sim}10$k epochs, App.~\ref{app:modadd-grokking}).  This speedup is absent for MoE-GLU. Thus, learned routing has measurable effects, but primarily on \emph{when} grokking occurs rather than on the learned attention--FFN division of labor.

% This dissociation strengthens the structural interpretation: redistribution responds to the sparse FFN bottleneck, not to learned router specialization.

% \paragraph{Why this matters.}
% A common reading of MoE is that the router learns task-specific specialization that explains the architectural advantages. In our controlled setting, this picture is incomplete: at parameter parity, sparse partitioning of FFN capacity (learned or otherwise) forces attention to compensate during training. This places at least part of the redistribution in the same broad category as other sparsity-driven training phenomena, such as dropout-induced regularization \citep{srivastava2014dropout}, rather than requiring the router itself to be a load-bearing learned component. We hypothesize the redistribution reflects an optimization asymmetry rather than a capacity limit: with per-token FFN capacity reduced, the loss gradient pulls substitutable computation into attention during training, even when both components could support the task under frozen-component training. The random-routing equivalence supports this reading for the add-7 ablation-visible effect: attention is reacting to the structural fact of the bottleneck, not to a co-adapting router.

\paragraph{Why this matters.}
A common interpretation of MoE is that learned routers induce task-specific expert specialization. Our controls show that this explanation is incomplete for the redistribution effect studied here. The narrow-FFN control accounts for part of the FFN-to-MoE gap through reduced per-token capacity, while random routing preserves the remaining effect without learned router adaptation. Thus, at parameter parity, sparse expert partitioning itself is sufficient to shift substitutable computation toward attention. In this setting, the router need not be a load-bearing learned mechanism for redistribution to occur. This points to an optimization-side interpretation, specifically when the FFN path is sparsely bottlenecked, training reallocates more substitutable work to attention.

%App.~\ref{app:frozen} shows MoE's FFN reaches $100\%$ on add-7 when attention is frozen).

%-------------------------------------------------------------------
\subsection{GLU Rotates Structure Out of the Neuron Basis}
\label{sec:glu}

\begin{figure}[htbp]
  \centering
  \includegraphics[width=0.65\linewidth]{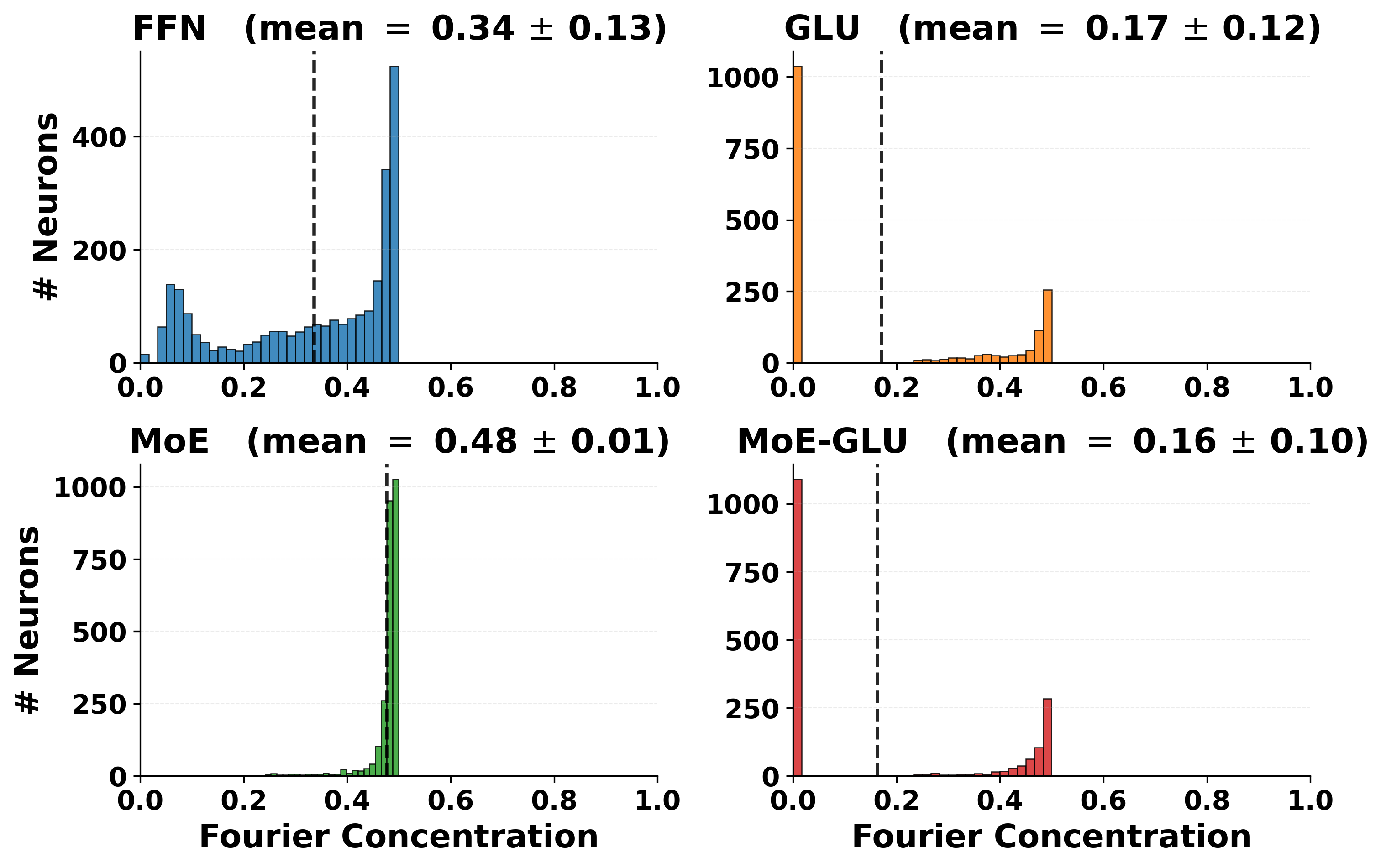}
  \caption{\textbf{GLU reshapes neuron-level Fourier structure.} Per-neuron Fourier concentration on $(a+b)\bmod 113$ at total-parameter-matched widths, pooled across 5 seeds. Mean per-neuron concentration $\pm$ seed-mean std: FFN $0.34 \pm 0.13$, MoE $0.48 \pm 0.01$, GLU $0.17 \pm 0.12$, MoE-GLU $0.16 \pm 0.10$. The gap between MoE-class and GLU-class variants is robust across seed-to-seed variation. All variants reach $\sim 100\%$ test accuracy. Full neuron counts and per-PC breakdown in App.~\ref{app:glu-pca}.}
  \label{fig:fourier-concentration}
\end{figure}

The GLU result is qualitatively different from the MoE result: GLU does not move computation into attention, but it changes the coordinate system in which FFN computation is represented. Unlike MoE, GLU changes the FFN nonlinearity without introducing sparse routing. Parameter-matched GLU models collapse to the dense-FFN baseline under FFN ablation, ruling out multiplicative gating as the source of the MoE redistribution effect.

On modular addition, FFNs exhibit the expected neuron-level Fourier structure. In contrast, GLU substantially reduces per-neuron Fourier concentration despite reaching the same task accuracy. This does not mean the algorithmic structure is absent. Linear probes recover the task information from the full activation vector, and PCA reveals Fourier-aligned directions in a low-dimensional subspace. GLU therefore preserves the computation while rotating it out of the neuron basis.

%%%%%%%%%%%%%%%%%%%%%%%%%%%%%%%%%%%%%%%%%%%%%%%%%%%%%%%%%%%%
\section{Discussion \& Conclusion}

\paragraph{Summary.}
In the setting we study, changing the FFN architecture is not a local modification. MoE's sparse routing bottleneck reduces active FFN capacity per token, and the trained model compensates by placing more substitutable computation in attention, as measured by higher accuracy when the FFN output is ablated. The effect is largest when the dense-baseline FFN work is both load-bearing and substitutable by attention. A narrow-FFN control attributes roughly one third of the FFN-to-MoE gap to reduced per-token capacity, while the remaining gap is associated with sparse partitioning across experts. A random-routing control with router weights frozen at initialization preserves comparable no-FFN accuracy to learned routing, showing that this partitioning component does not require learned router adaptation. Top-2 routing confirms the bottleneck interpretation, when each token can access more expert capacity, the ablation pattern moves back toward the dense baseline.

These results support an optimization-side interpretation rather than a simple capacity-limit interpretation. Frozen-component training shows that either component can support the task when the other is fixed, but joint training with a sparse FFN bottleneck changes where the learned model places the computation. We conjecture that, when active FFN capacity is reduced, gradient descent reallocates substitutable work into attention. In this sense, the redistribution resembles other sparsity-driven training phenomena, such as dropout-induced regularization \citep{srivastava2014dropout}, more than it resembles a mechanism that requires the router itself to become a load-bearing learned component.

Histogram counting marks the boundary of this effect. Sparse partitioning can shift internal strategy, but frozen-component solvability does not imply ablation substitutability. Although histogram can be learned when either attention or the FFN is frozen but still present (App.~\ref{app:frozen}), jointly trained models still collapse when the FFN output is removed, indicating complementary rather than ablation-substitutable attention--FFN roles. GLU provides a different contrast, its multiplicative gating does not produce the same FFN-to-attention redistribution, but rotates Fourier structure out of the per-neuron basis into distributed subspaces, preserving task information while making it less visible to standard neuron-level tools.

\paragraph{Scope and limitations.}
Our experiments use one-layer Transformers and small algorithmic tasks by design. This setting allows component ablation, direct logit attribution, and activation patching to have relatively direct interpretations. We do not claim that the quantitative redistribution effects reported here transfer directly to large language models. Deeper models introduce additional forms of redistribution across layers, residual streams, and attention heads that are not captured in this study. Our claim is instead mechanistic: in a controlled setting, sparse FFN partitioning is sufficient to change what attention learns to compute, even without learned routing. Whether analogous pressures appear in deeper MoE language models is an important direction for future work.

A second limitation is that component ablation measures functional reliance after training, not the entire causal computation of the unablated model. We therefore interpret ablation together with direct logit attribution, activation patching, frozen-component controls, and task-level negative controls. Finally, the magnitude of the redistribution depends on the matching convention. Total-parameter matching exposes the sparse per-token bottleneck most directly, while FLOP-matched controls show that the direction persists but the effect is smaller.

\paragraph{Reproducibility.}
All code, checkpoints, and analysis scripts are available at \url{https://github.com/dipplestix/moe-expressivity}. All experiments were run on a single MacBook Pro (Apple M1 Pro, 8 CPU cores, 16 GB unified memory); no external GPUs, accelerators, or cloud compute were used. Models were trained on CPU/MPS with PyTorch.                                               

% \paragraph{Reproducibility.}
% All code, checkpoints, and analysis scripts are available at \url{https://anonymous.4open.science/r/sparsity-moves-computation-4440/README.md}. All experiments were run on a single MacBook Pro (Apple M1 Pro, 8 CPU cores, 16 GB unified memory); no external GPUs, accelerators, or cloud compute were used. Models were trained on CPU/MPS with PyTorch.                                               

%%%%%%%%%%%%%%%%%%%%%%%%%%%%%%%%%%%%%%%%%%%%%%%%%%%%%%%%%%%%

\bibliographystyle{plainnat}
\bibliography{references, references2}

\newpage
\appendix
\section{Experimental Setup Details}
\label{app:setup-details}

\paragraph{Add-7 task.}
Given a number $x$, compute $x + 7$ digit by digit. Numbers are in reversed order (ones first). The input format is $[d_0, d_1, \ldots, d_n, \text{EOS}, o_0, o_1, \ldots, o_{n+1}, \text{EOS}]$. Example: $593 + 7 = 600$ becomes input $[3, 9, 5, \text{EOS}]$, output $[0, 0, 6, 0, \text{EOS}]$. Each output position requires $+7$ (ones), $+1$ (carry), or $+0$ (pass-through). Carry-chain length $L$ counts consecutive carries: $L{=}0$ (no overflow, e.g.\ $122 \to 129$), $L{=}1$ (one carry, e.g.\ $193 \to 200$), $L{=}2$ (two carries, e.g.\ $993 \to 1000$). We use 3-digit numbers (000-999) with 4 output positions. Dataset: 1000 $+7$ tokens, 777 $+1$ tokens, 2223 $+0$ tokens.

\paragraph{Modular addition.}
Given $a, b \in \mathbb{Z}_{113}$, predict $(a + b) \bmod 113$. Input: $[a, b, \mathord{=}]$, vocabulary size 114. Loss at position 2. Train/test: 30\%/70\% of all $113^2$ pairs. Full-batch training, 40k epochs, weight decay 1.0. Networks develop discrete Fourier circuits with per-neuron frequency selectivity \citep{nanda2023progress}.

\paragraph{Histogram counting.}
Given $L{=}10$ tokens from alphabet $T{=}32$, predict each token's count. Illustrative $L{=}6$ example (the full task uses $L{=}10$): $[2, 0, 3, 3, 0, 0] \to [1, 3, 2, 2, 3, 3]$ (token $2$ appears once, token $0$ appears three times, token $3$ appears twice). Backward sampling ensures uniform partition structures \citep{Behrens2024CountingIS}. Bidirectional attention. Full-batch, 1000 epochs, data refresh every 50 epochs.

% \paragraph{Architecture.}
% Add-7: $d_{\text{model}}{=}64$, 4 heads, tied embeddings. FFN hidden dim $256$; GLU hidden dim $h{=}170$; MoE with $E{=}4$ experts each at $h_E{=}64$; MoE-GLU with $E{=}4$ experts each at $h_E{=}42$ (total expert width $4\cdot 42{=}168 \approx \tfrac{2}{3}h_{\text{dense}}$). Modular addition: $d_{\text{model}}{=}128$, 4 heads, untied output head. FFN hidden dim $512$; GLU $h{=}340$; MoE $h_E{=}128$; MoE-GLU $h_E{=}85$ (total expert width $340$). Histogram: $d_{\text{model}}{=}128$, 4 heads, bidirectional attention, untied output head (output classes are token counts $0,\ldots,L$, distinct from the input vocabulary); same FFN widths as modular addition. Linear layers in FFN and MoE include biases; GLU and MoE-GLU layers omit biases, following the standard GLU formulation, which is reflected in the parameter counts of Tab.~\ref{tab:ffn-params}. All four FFN variants are total-parameter-matched within 1-2\% (Tab.~\ref{tab:ffn-params}). MoE variants use top-1 routing with auxiliary load-balancing loss ($\lambda_{\text{aux}}{=}0.01$) for all main results; top-2 routing is reported as a control (App.~\ref{app:topk}) and uses the same architecture (same $E$, same $h_E$) with $k{=}2$ routing, isolating the routing-$k$ axis at fixed architecture. No RMSNorm for primary analyses. 

\paragraph{Architecture.}
\label{app:architecture-details}
Add-7 uses $d_{\text{model}}{=}64$, 4 heads, and tied embeddings. Its FFN hidden dimension is $256$; GLU uses hidden dimension $h_{\text{glu}}{=}170$; MoE uses $E{=}4$ experts each with $h_E{=}64$; and MoE-GLU uses $E{=}4$ experts each with $h_E{=}42$ (total expert width $4\cdot 42{=}168 \approx \tfrac{2}{3}h_{\text{dense}}$). Modular addition uses $d_{\text{model}}{=}128$, 4 heads, and an untied output head. Its FFN hidden dimension is $512$; GLU uses $h_{\text{glu}}{=}340$; MoE uses $h_E{=}128$; and MoE-GLU uses $h_E{=}85$ per expert (total expert width $340$). Histogram uses $d_{\text{model}}{=}128$, 4 heads, bidirectional attention, and an untied output head because the output classes are token counts $0,\ldots,L$, distinct from the input vocabulary; it uses the same FFN widths as modular addition.

Linear layers in FFN and MoE include biases; GLU and MoE-GLU layers omit biases, following the standard GLU formulation, which is reflected in the parameter counts of Tab.~\ref{tab:ffn-params}. All four FFN variants are total-parameter-matched within $1$--$2\%$ (Tab.~\ref{tab:ffn-params}). Concretely, $h_{\text{glu}}{=}170$ for $d{=}64$ ($32{,}640$ vs.\ FFN's $33{,}088$ parameters) and $h_{\text{glu}}{=}340$ for $d{=}128$ ($130{,}560$ vs.\ FFN's $131{,}712$ parameters). MoE-GLU is total-parameter-matched-as-GLU: total expert width is $\sum_e h_E = \tfrac{2}{3}h_{\text{dense}}$, with each expert at $h_E = \tfrac{2}{3}h_{\text{dense}}/E$ ($h_E{=}42$ on add-7, $h_E{=}85$ on modular addition and histogram). Total MoE-GLU FFN parameters match FFN within $1$--$2\%$, and per-token active capacity is $1/E$ of dense.

MoE variants use top-1 routing with auxiliary load-balancing loss ($\lambda_{\text{aux}}{=}0.01$) for all main results. The redistribution effect we report is therefore measured under total-parameter matching: MoE variants have no more total FFN parameters than the dense baseline, while each token accesses strictly fewer active FFN parameters (Tab.~\ref{tab:ffn-params}). Top-2 routing is reported as a control (App.~\ref{app:topk}) and uses an architecturally identical model with the same $E$ and same $h_E$, but with $k{=}2$ routing, isolating the routing-$k$ axis at fixed architecture. We train without RMSNorm for primary analyses. All experiments use 5 seeds. Full task descriptions, architecture details, and analysis method definitions are in App.~\ref{app:setup-details}.

\begin{table}[h]
\centering
\small
\begin{tabular}{lcccc}
\toprule
\textbf{Task} ($d_m$) & \textbf{FFN} & \textbf{GLU} & \textbf{MoE} & \textbf{MoE-GLU} \\
& ($h$, params) & ($h$, params) & ($h_E$, params) & ($h_E$, params) \\
\midrule
Add-7 ($d_m{=}64$)              & $h{=}256$, $33{,}088$ & $h{=}170$, $32{,}640$ & $h_E{=}64$, $33{,}536$ & $h_E{=}42$, $\phantom{0}32{,}512$ \\
Modular addition ($d_m{=}128$) & $h{=}512$, $131{,}712$ & $h{=}340$, $130{,}560$ & $h_E{=}128$, $132{,}608$ & $h_E{=}85$, $131{,}072$ \\
Histogram ($d_m{=}128$)        & $h{=}512$, $131{,}712$ & $h{=}340$, $130{,}560$ & $h_E{=}128$, $132{,}608$ & $h_E{=}85$, $131{,}072$ \\
\bottomrule
\end{tabular}
\caption{FFN parameter counts at the matched widths used in our experiments. All four variants are total-parameter-matched to FFN within 1-2\%. GLU uses $h{=}\tfrac{2}{3}h_{\text{dense}}$ to absorb its third weight matrix. MoE uses $h_E{=}h_{\text{dense}}/E$ per expert. MoE-GLU uses $h_E{=}\tfrac{2}{3}h_{\text{dense}}/E$, so total expert width sums to $\tfrac{2}{3}h_{\text{dense}}$ and total parameters match FFN, while per-active capacity is $1/E$ of FFN. App.~\ref{app:param-vs-flop} discusses the trade-off vs.\ per-active (FLOP) matching, and App.~\ref{app:per-active-controls} reports a per-active-matched MoE-GLU control.}
\label{tab:ffn-params}
\end{table}

\begin{table}[h]
\centering
\small
\begin{tabular}{lccc}
\toprule
& \textbf{Add-7} & \textbf{Modular Addition} & \textbf{Histogram} \\
\midrule
Optimizer & AdamW & AdamW & AdamW \\
Learning rate & $10^{-3}$ & $10^{-3}$ & $10^{-3}$ \\
Weight decay & 0 & 1.0 & 0.1 \\
Batch size & 128 (online) & full-batch & full-batch \\
Duration & 10k steps & 40k epochs & 1000 epochs \\
Train/test & online & 30\%/70\% & 10k/3k \\
\bottomrule
\end{tabular}
\caption{Training hyperparameters. All experiments use 5 seeds (42, 137, 256, 512, 1024).}
\label{tab:hyperparams}
\end{table}

\section{Analysis Method Details}
\label{app:analysis-details}

\begin{figure}[htbp]
  \centering
  \includegraphics[width=0.85\linewidth]{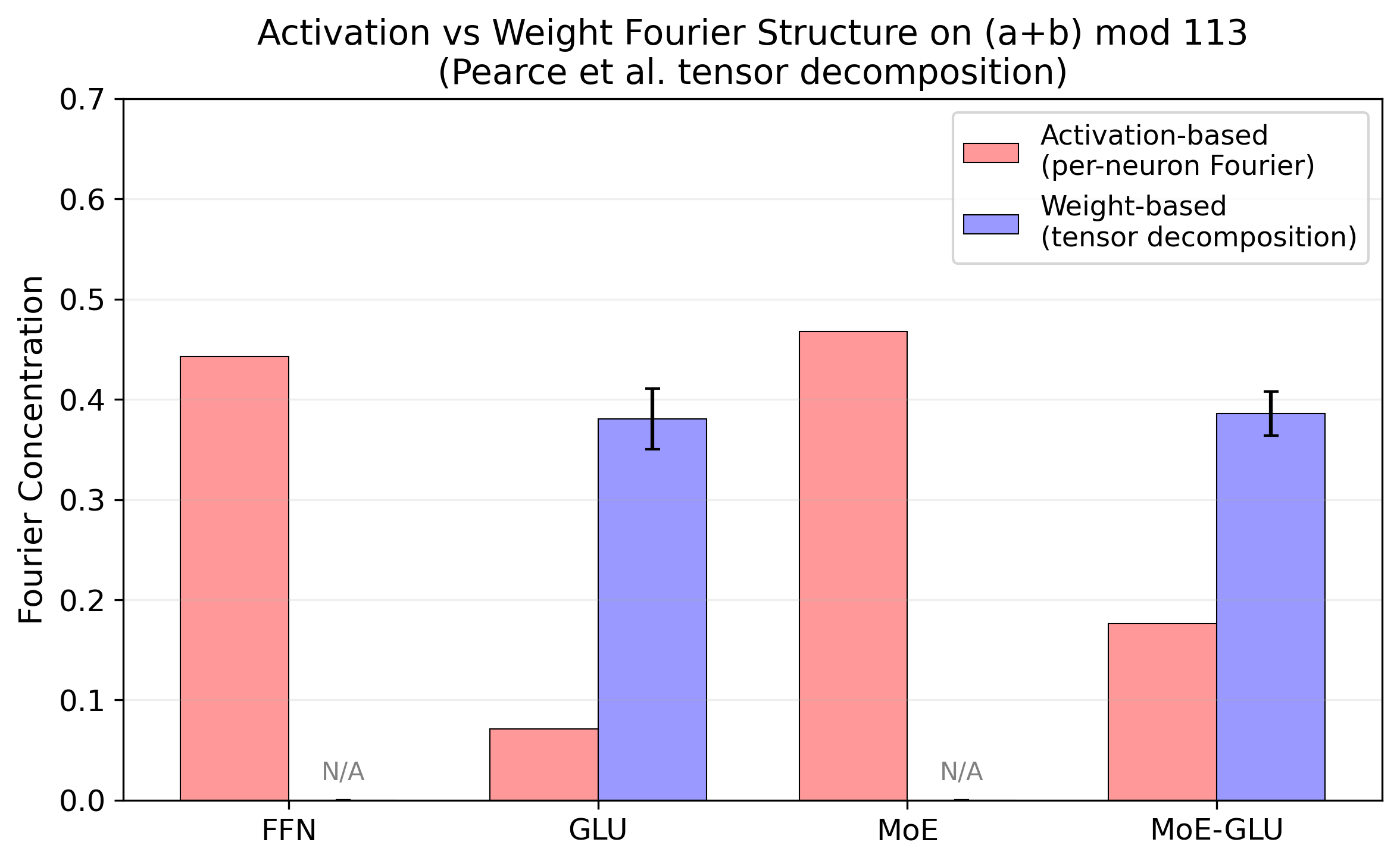}
  \caption{Weight vs.\ activation Fourier concentration for GLU and MoE-GLU on $(a{+}b)\bmod 113$. The weight-based metric applies the bilinear tensor of \citet{pearce2024bilinear} with the singular vectors projected into digit-token space via the input embeddings (full pipeline in App.~\ref{app:analysis-details},
  item~5b). Weight-based concentrations ($0.38 \pm 0.03$ for GLU, $0.39 \pm 0.02$ for MoE-GLU; 5 seeds) are substantially higher than per-neuron activation concentrations
   on the same checkpoints ($0.07$ and $0.18$). The dissociation directly demonstrates the rotated-subspace interpretation: Fourier structure is preserved in the GLU
  weight bilinear tensor when viewed in token-space coordinates, but is rotated out of the per-neuron activation basis at inference time. FFN and plain MoE are not       
  bilinear, so the weight-based metric is undefined (N/A); their activation-based bars are reported for comparison.}
  \label{fig:glu-weight-activation}
\end{figure}

\begin{enumerate}
\item \textbf{Component ablation}: Zero attention or FFN output at inference and measure per-token output accuracy degradation. This reveals each component's contribution to the final prediction.
\item \textbf{Direct logit attribution (DLA)}: Decompose output logits into additive contributions via the residual stream \citep{nanda2023progress, quirke2023understanding}: $\text{logits} = W_U(x_{\text{embed}} + x_{\text{attn}} + x_{\text{ffn}})$. Each term's projection through $W_U$ gives that component's contribution to the correct output logit.
\item \textbf{Activation patching}: For pairs of inputs requiring different operations at the same position, swap attention or FFN activations and measure whether the prediction flips. This provides \emph{causal} evidence for which component carries the decision \citep{meng2022locating}.
\item \textbf{Linear probes}: Train a single layer (logistic regression) to predict the operation type ($+7$/$+1$/$+0$) from attention or FFN outputs. Trained with Adam ($\text{lr}{=}10^{-3}$, no weight decay), cross-entropy loss, batch size 128, for 500 steps on a 70/30 train/test split of the 1000 examples. This is the weakest possible probe, so 100\% accuracy means the features are linearly separable in the activation space.
\item \textbf{Fourier analysis (activation-based)}: For modular addition, compute per-neuron Fourier concentration as the fraction of spectral power at the dominant frequency when activations are viewed as a function of $(a + b) \bmod p$ \citep{nanda2023progress}. High concentration means the neuron responds cleanly to one frequency. For add-7, where there is no canonical cyclic group, we apply the same protocol with the input grouping replaced by the input number $n \in \{0, \ldots, 999\}$ taken mod $10$ at each digit position, treating the per-position digit cycle as the analogue of $\mathbb{Z}_{10}$. The add-7 Fourier rows (Tab.~\ref{tab:fourier-neuron-control}, top) report this protocol.
\item \textbf{Fourier analysis (weight-based, GLU only)}: Used in Fig.~\ref{fig:glu-weight-activation} as a weight-space analogue of the activation-based metric above. The pipeline:
\begin{enumerate}
\item \emph{Bilinear tensor.} Following the bilinear interpretation of GLU \citep{pearce2024bilinear}, form the third-order tensor
$T_{ijk} = \sum_m W_{\text{down}}[i,m]\, W_{\text{gate}}[m,j]\, W_{\text{up}}[m,k]$
from the raw weight matrices, so that $\text{GLU}(x)_i \approx \sum_{j,k} T_{ijk}\, x_j\, x_k$ (treating the gate input and up-projection input as the same $x$). This is exact for bilinear MLPs (where $\sigma$ is the identity); for SiLU/GELU GLU it drops the gate activation, so $T$ is the bilinear approximation of the full GLU function. 

%We treat the resulting concentration as ``what would be visible to a Pearce-style weight-only analyst'' rather than an exact characterization of GLU's input-output behaviour.

\item \emph{SVD of the unfolding.} Reshape $T$ to its mode-1 unfolding $\tilde T \in \mathbb{R}^{d_{\text{out}}\times d_{\text{in}}^2}$ and take SVD. Each right singular vector $v_k$ reshapes to a $d_{\text{in}}\times d_{\text{in}}$ matrix $V_k$ capturing a dominant bilinear interaction pattern in residual-stream coordinates.
\item \emph{Project into digit-token space.} Since $V_k$ lives in the residual-stream basis, we first evaluate it on the embedding representations of the valid digit tokens. Let $E_p \in \mathbb{R}^{p\times d_{\text{in}}}$ contain the input residual embeddings for digit tokens $0,\ldots,p-1$. We form the token-space interaction matrix             
\[    
M_k = E_p\, V_k\, E_p^\top \in \mathbb{R}^{p\times p}.            
\]                                       We then compute the modular-diagonal average              
\[                                       
\bar M_k(s) = \frac{1}{p}\sum_{a=0}^{p-1} M_k[a,\,(s-a)\bmod p],\qquad s \in \{0,\ldots,p-1\}.                         \]                              
This gives a length-$p$ signal per singular vector, defined over valid digit-token pairs rather than residual-stream coordinates.        
\item \emph{Fourier concentration.} Compute $\text{conc}(v_k) = \max_{f\neq 0} S_f / \sum_{f\neq 0} S_f$ on $\bar M_k$ (DC dropped), where $S_f = |\hat{\bar M}_k(f)|^2$.
\item \emph{Aggregate.} Mean over the top-10 right singular vectors per checkpoint; for MoE-GLU first within each expert, then across experts; finally across 5 seeds.
\end{enumerate}
The token-space projection makes the rotated-subspace interpretation directly testable. If the GLU's Fourier structure is rotated out of the per-neuron activation basis but preserved in the weight bilinear tensor, weight-based concentration on $(a+b)\bmod p$ should be high while activation-based per-neuron concentration is low.
Fig.~\ref{fig:glu-weight-activation} shows exactly this dissociation: weight-based concentration is $0.38 \pm 0.03$ for GLU and $0.39 \pm 0.02$ for MoE-GLU (5 seeds), while the per-neuron activation concentrations on the same checkpoints are $0.07$ and $0.18$ respectively. The structure is preserved in the GLU weights when viewed in
token coordinates and is rotated out of the per-neuron activation basis at inference time.
\item \textbf{Routing analysis} (MoE): Compute normalized mutual information between expert assignment and operation type. Ablate individual experts and measure per-operation accuracy drops.
\end{enumerate}

%% ---- C: Redistribution Supporting Evidence ----
\section{Redistribution Supporting Evidence}
\label{app:redistribution}

\subsection{Norm vs.\ No-Norm Comparison}
\label{app:norm}

\begin{figure}[htbp]
  \centering
  \begin{subfigure}[t]{0.48\linewidth}
    \centering
    \includegraphics[width=\linewidth]{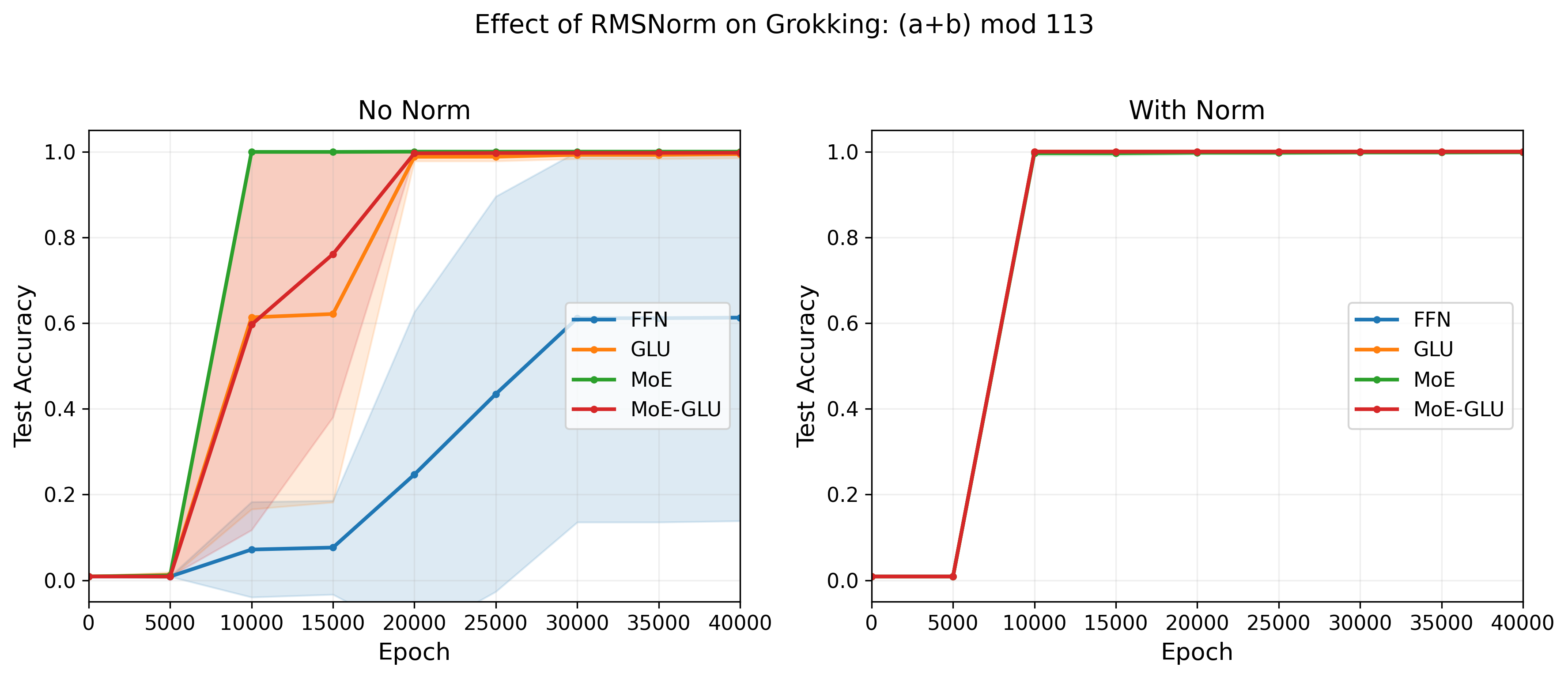}
    \caption{Norm vs.\ no-norm grokking.}
  \end{subfigure}
  \hfill
  \begin{subfigure}[t]{0.48\linewidth}
    \centering
    \includegraphics[width=\linewidth]{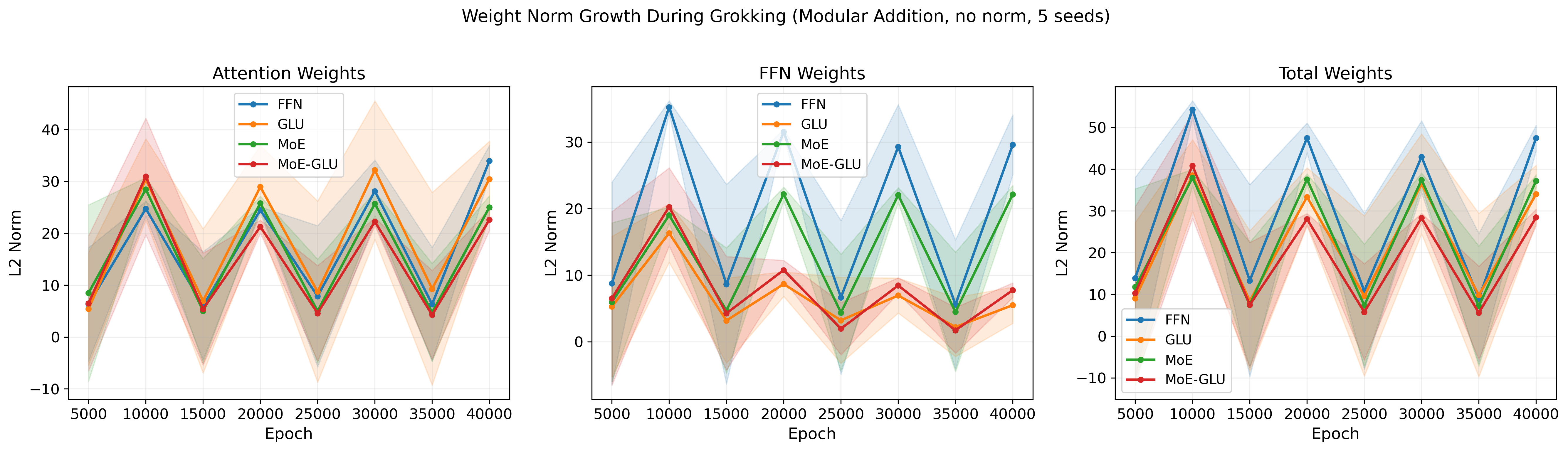}
    \caption{Weight norm dynamics. MoE compresses norms $0.63\times$. FFN attention norms grow $1.35\times$. GLU FFN norms collapse to $0.21\times$.}
  \end{subfigure}
  \caption{Norm comparisons and weight norm dynamics during grokking.}
  \label{fig:norm-weight}
\end{figure}

We train without RMSNorm following standard practice in mechanistic interpretability \citep{nanda2023progress, quirke2023understanding}. All main results use no-norm checkpoints. With RMSNorm, qualitative patterns persist but quantitative differences compress (Fig.~\ref{fig:norm-weight}).

\begin{figure}[htbp]
  \centering
  \includegraphics[width=0.85\linewidth]{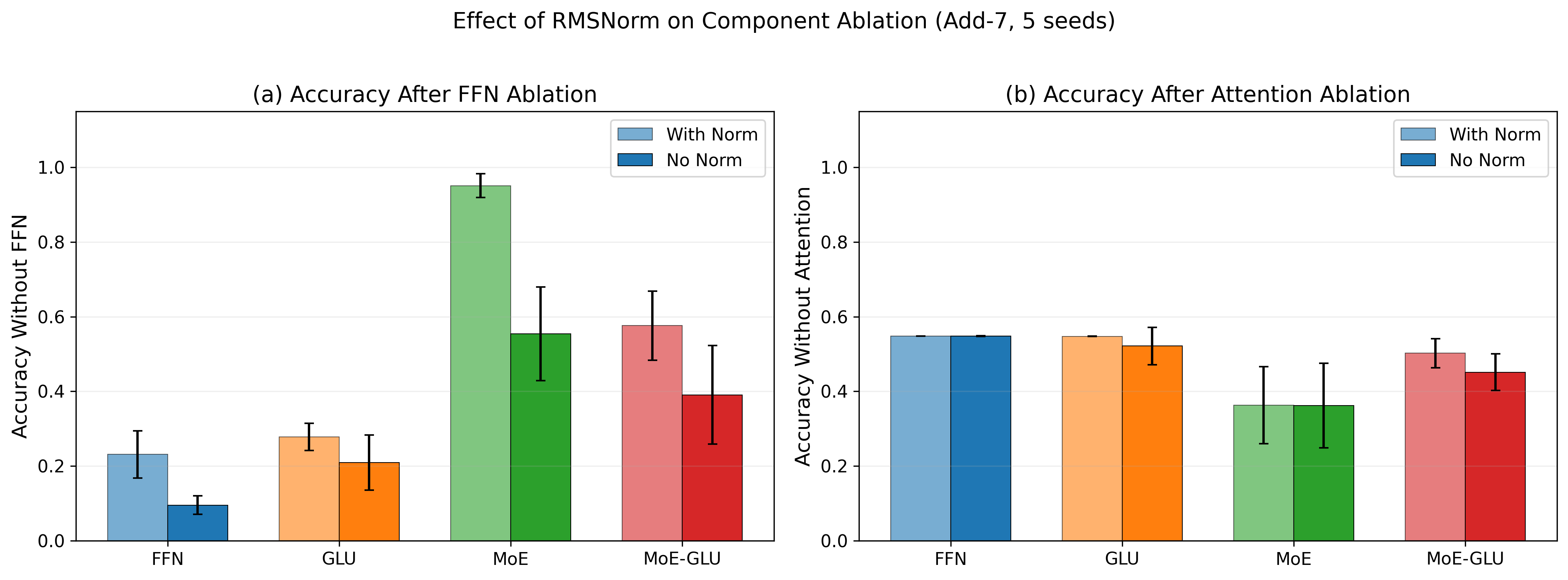}
  \caption{Norm effect on component ablation (add-7). No-norm produces cleaner separation.}
  \label{fig:norm-ablation}
\end{figure}

\FloatBarrier
\subsection{Direct Logit Attribution}
\label{app:dla}

\begin{figure}[htbp]
  \centering
  \includegraphics[width=0.85\linewidth]{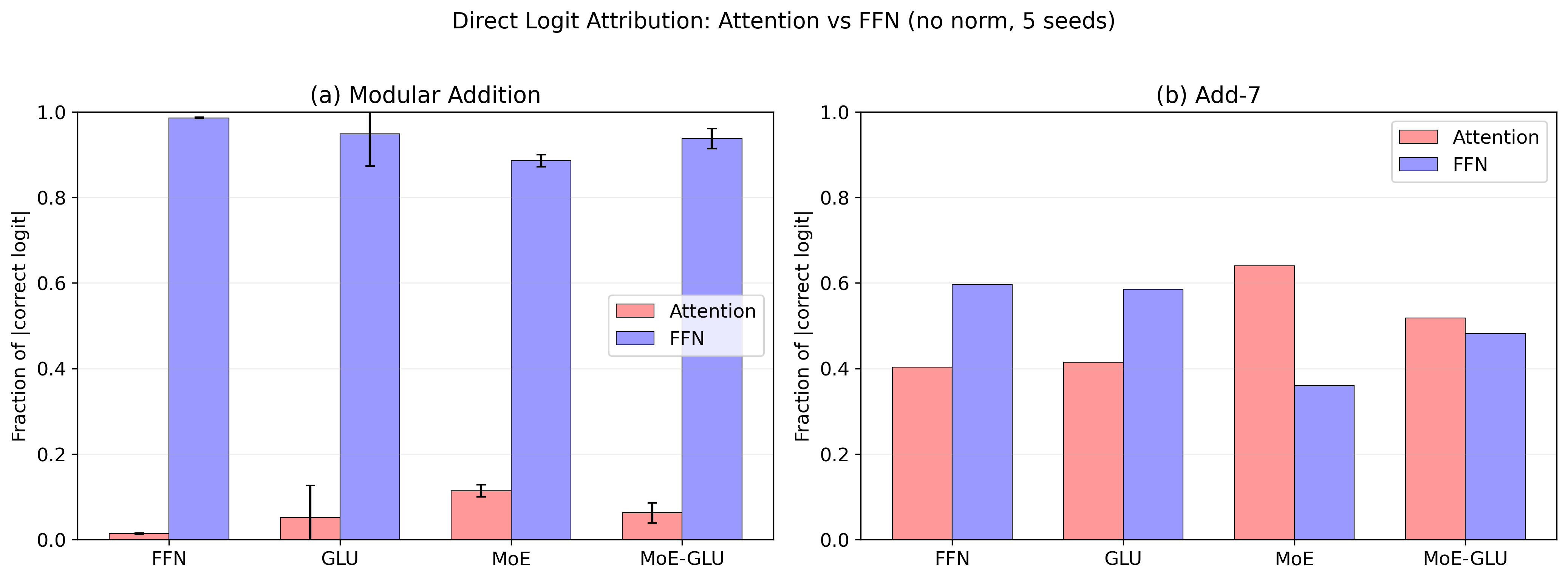}
  \caption{\textbf{Direct logit attribution across tasks} (no norm, 5 seeds). Each bar shows the fraction of the correct output logit attributable to attention vs.\ FFN. \textbf{(a)} On modular addition, FFN dominates (88-99\%). \textbf{(b)} On add-7, MoE's attention contributes ${\sim}60\%$ vs.\ ${\sim}40\%$ for FFN. Error bars: $\pm 1$ std across 5 seeds.}
  \label{fig:dla-cross-task}
\end{figure}

\begin{figure}[htbp]
  \centering
  \includegraphics[width=0.85\linewidth]{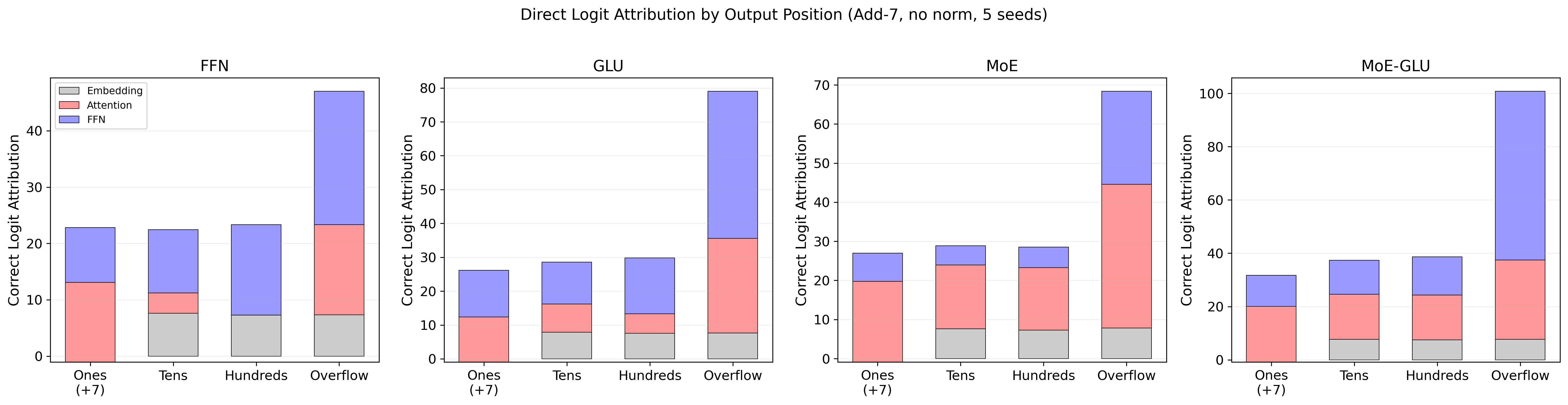}
  \caption{DLA by output position on add-7 (5 seeds). Error bars: $\pm 1$ std.}
  \label{fig:dla-by-position}
\end{figure}

\begin{figure}[htbp]
  \centering
  \includegraphics[width=0.6\linewidth]{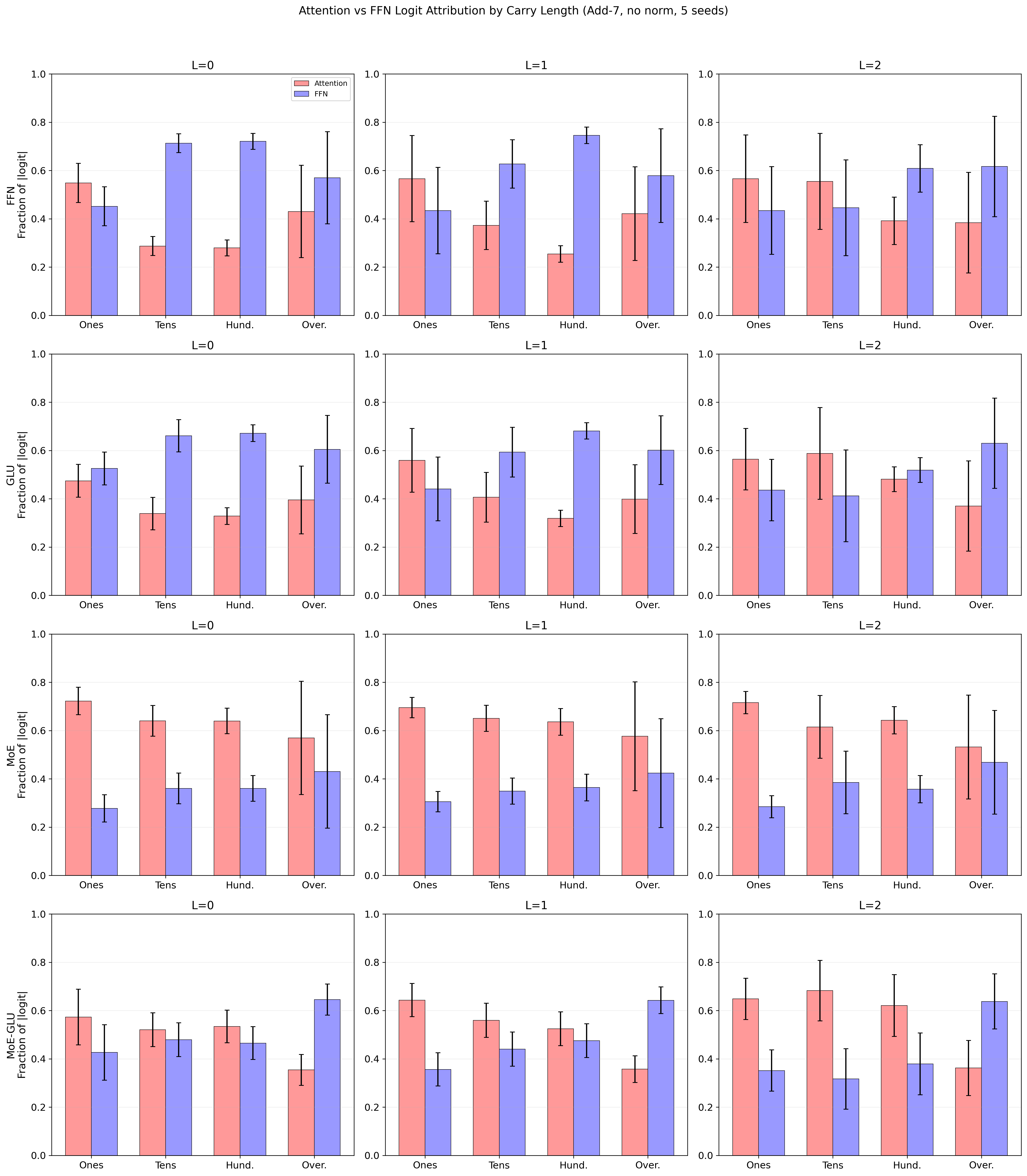}
  \caption{DLA by carry-chain length on add-7 (5 seeds). Each row is a variant; columns are $L{=}0,1,2$. Error bars: $\pm 1$ std.}
  \label{fig:dla-by-carry}
\end{figure}

\subsubsection{Per-head DLA}
\label{app:per-head-dla}

The aggregate DLA results show that MoE increases the attention-side contribution to the correct output logit. We further ask whether this increase is diffuse across attention heads or concentrated in a small number of heads. Head identities are permutation-arbitrary across random seeds, we sort heads within each seed by total absolute correct-logit contribution before averaging. This rank-based comparison avoids treating, for example, Head~0 in different seeds as the same circuit. 

% Figure~\ref{fig:dla-per-head} shows that the head-level structure mirrors the causal decomposition in Sec.~\ref{sec:routing-causal}. At carry-propagation positions, the FFN baseline is close to a uniform split across heads, with carry-position-mean Rank-1 share $0.29$ compared to the uniform value $1/H=0.25$. Reducing active FFN capacity increases concentration: narrow FFN reaches Rank-1 share $0.57$. Sparse partitioning increases it further: random-routing MoE reaches $0.63$, matching learned MoE at $0.63$. Thus, the dominant-head contribution does not require learned routing; as with the aggregate redistribution result, it is explained by reduced active capacity and sparse partitioning.

\begin{figure}[htbp]
\centering
\includegraphics[width=0.95\linewidth]{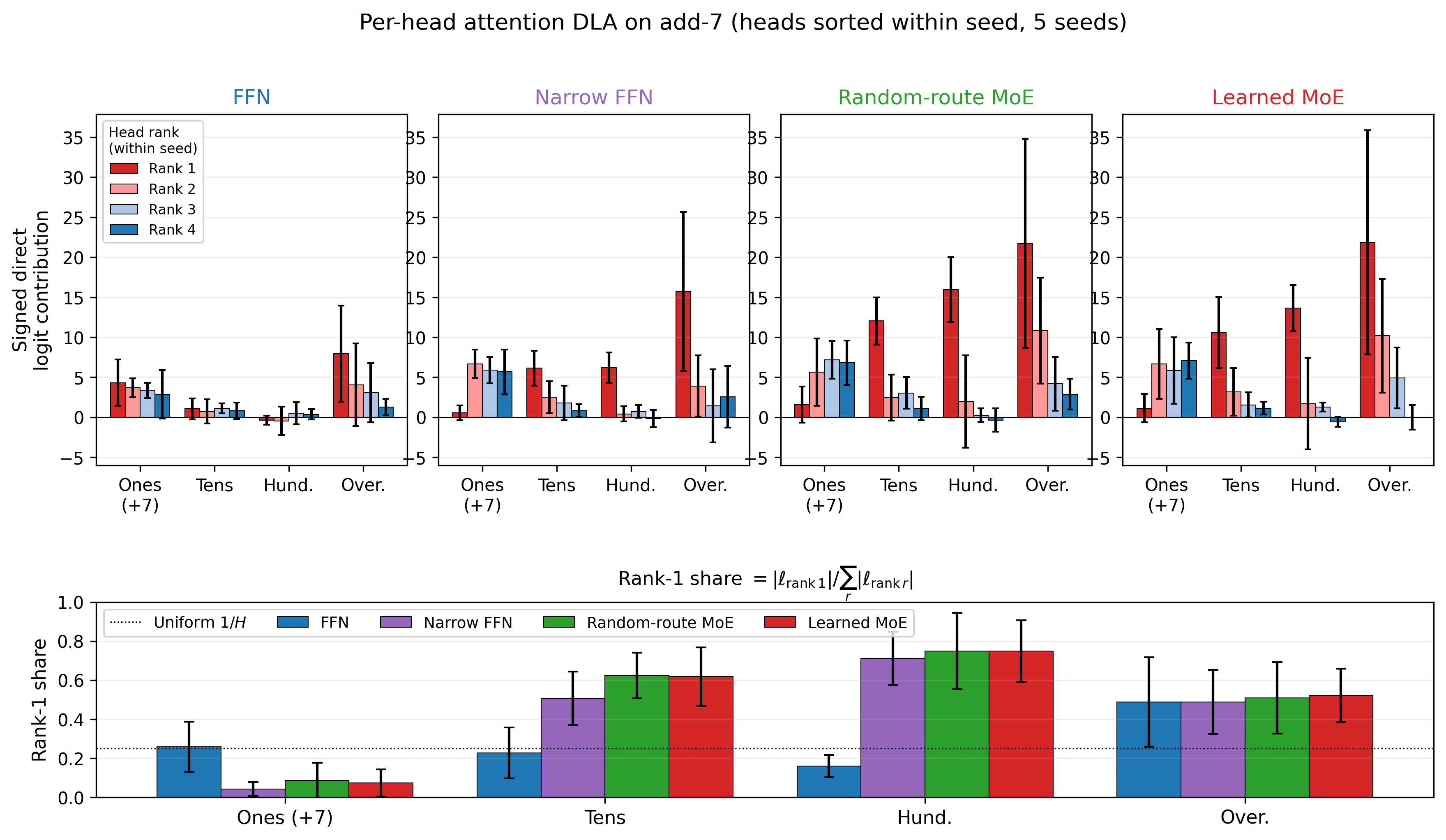}
\caption{\textbf{Head-level attention concentration tracks the FFN capacity bottleneck.}
Per-head direct logit attribution on add-7, split by output position. Heads are sorted within each seed by total absolute correct-logit contribution before averaging, so Rank~1 is the largest contributor within that seed rather than a fixed head index. The four conditions match the causal decomposition in Sec.~\ref{sec:routing-causal}: FFN, narrow FFN, random-routing MoE, and learned MoE. \textbf{Top:} signed per-head correct-logit contribution by output position. \textbf{Bottom:} Rank-1 share, $|\ell_{\mathrm{rank}\,1}|/\sum_{r=1}^{H}|\ell_{\mathrm{rank}\,r}|$, with $1/H{=}0.25$ corresponding to a uniform split across heads. At the intermediate carry positions (Tens and Hundreds), the FFN baseline is close to uniform (Rank-1 share $\approx 0.29$ on average), while narrow FFN ($0.57$), random-routing MoE ($0.63$), and learned MoE ($0.63$) increasingly concentrate attention-side contribution in the top-ranked head. The Overflow position is already strongly concentrated in the dense FFN baseline, and the restricted-capacity variants do not raise it further, dominant-head shift is an effect at the intermediate carry positions rather than uniformly across carry-propagation positions. Random-routing MoE matches learned MoE in Rank-1 share and top-head magnitude, indicating that the dominant-head contribution does not require learned routing. Error bars show $\pm 1$ s.d.\ across five seeds. The shared $o_\text{proj}$ bias is excluded from the per-head decomposition.}
\label{fig:dla-per-head}
\end{figure}

Figure~\ref{fig:dla-per-head} shows that the head-level structure mirrors the causal decomposition in Sec.~\ref{sec:routing-causal}, but the trend is driven by the intermediate carry positions (Tens and Hundreds) rather than holding uniformly across all output positions. At Tens and Hundreds, the FFN baseline is close to a uniform split across heads (Rank-1 share $\approx 0.29$, compared to the uniform value $1/H=0.25$).  Reducing active FFN capacity raises this concentration to $0.57$ (narrow FFN), and sparse partitioning raises it further to $0.63$ for both random-routing and learned MoE. The Overflow position is different: the dense FFN baseline is already strongly concentrated at Overflow, so the restricted variants do not meaningfully raise it further. The dominant-head shift is therefore an effect at the intermediate carry positions, where redistribution is.

\FloatBarrier
\subsection{DLA vs.\ Activation Patching}
\label{app:dla-patching-note}

DLA and activation patching measure different quantities. DLA decomposes the logit for the correct token into additive contributions, measuring \emph{magnitude}. Patching measures whether swapping a component's output changes the discrete \emph{prediction} (argmax). For the FFN at the tens position, DLA shows FFN contributes ${\sim}60\%$ of the correct logit, but patching shows attention has a higher flip rate (${\sim}0.5$ vs.\ ${\sim}0.1$). This is consistent: attention carries the discriminative information between competing outputs even though FFN contributes more absolute magnitude. We report both methods as complementary rather than convergent.

\FloatBarrier
\subsection{Component Ablation Across All Tasks}
\label{app:ablation-summary}

\begin{table}[htbp]
\centering
\small
\begin{tabular}{lcccccc}
\toprule
& \multicolumn{3}{c}{\textbf{No Attention}} & \multicolumn{3}{c}{\textbf{No FFN}} \\
\cmidrule(lr){2-4} \cmidrule(lr){5-7}
\textbf{Variant} & Add-7 & ModAdd & Hist. & Add-7 & ModAdd & Hist. \\
\midrule
FFN     & 32.3\% & 0.88\% & 19.2\% & 11.9\% & 1.3\%  & 11.5\% \\
GLU     & 30.7\% & 0.88\% & 19.4\% & 9.7\%  & 1.4\%  & 9.8\%  \\
MoE     & 16.8\% & 0.88\% & 17.0\% & 44.3\% & 7.5\%  & 12.0\% \\
MoE-GLU & 21.8\% & 0.88\% & 18.6\% & 41.7\% & 2.6\%  & 10.5\% \\
\bottomrule
\end{tabular}
\caption{Component ablation summary at matched parameter widths (5 seeds, no norm). The uniform $0.88\%$ no-attention accuracy on modular addition (${=}1/113$, chance level) confirms that FFN alone cannot solve this task: it requires attention to route information between positions. The no-FFN column shows the redistribution effect.  At parity, MoE retains much higher no-FFN accuracy on tasks with ablation-visible redistribution pressure: largest on add-7 ($+32.4$\,pp over FFN), smaller but consistent on modular addition ($+6.2$\,pp), and not ablation-visible on histogram ($+0.5$\,pp). MoE-GLU follows the same pattern (${+}29.8$\,pp add-7). GLU at parity tracks FFN, isolating sparse routing as the redistribution mechanism.}
\label{tab:ablation-summary}
\end{table}

\begin{figure}[htbp]
  \centering
  \includegraphics[width=0.85\linewidth]{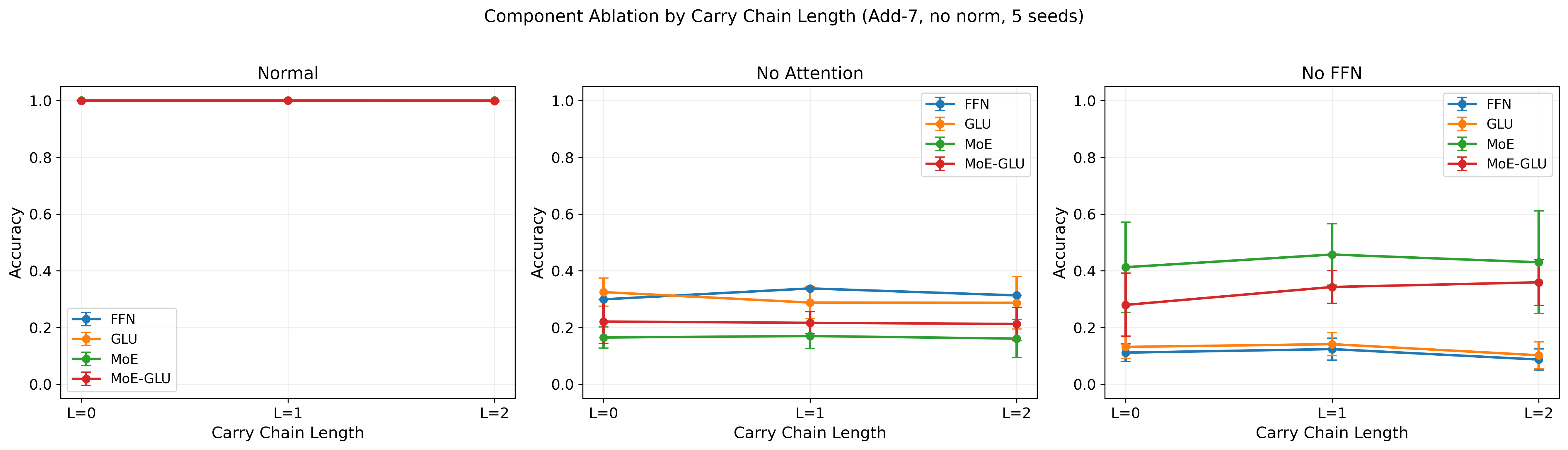}
  \caption{Ablation stratified by carry-chain length on add-7. MoE advantage under no-FFN ablation holds across all carry lengths. Error bars: $\pm 1$ std across 5 seeds.}
  \label{fig:ablation-carry}
\end{figure}

\FloatBarrier
\subsection{Attention Patterns and Activation Patching}
\label{app:attention-patching}

\paragraph{Activation patching methodology.}
For each output position $t \in \{\text{tens, hundreds, overflow}\}$, we construct matched pairs of inputs requiring different operations at position $t$. We use all valid pairs from the 1000-example test set (typically 200-400 pairs per position). For each pair, we run both inputs, capture attention and FFN outputs at position $t$, then re-run one input with the other's component output patched in. The flip rate is the fraction of pairs where patching changes the predicted token. Patching is one component and one position at a time; results averaged across 5 seeds.

\begin{figure}[htbp]
  \centering
  \includegraphics[width=0.85\linewidth]{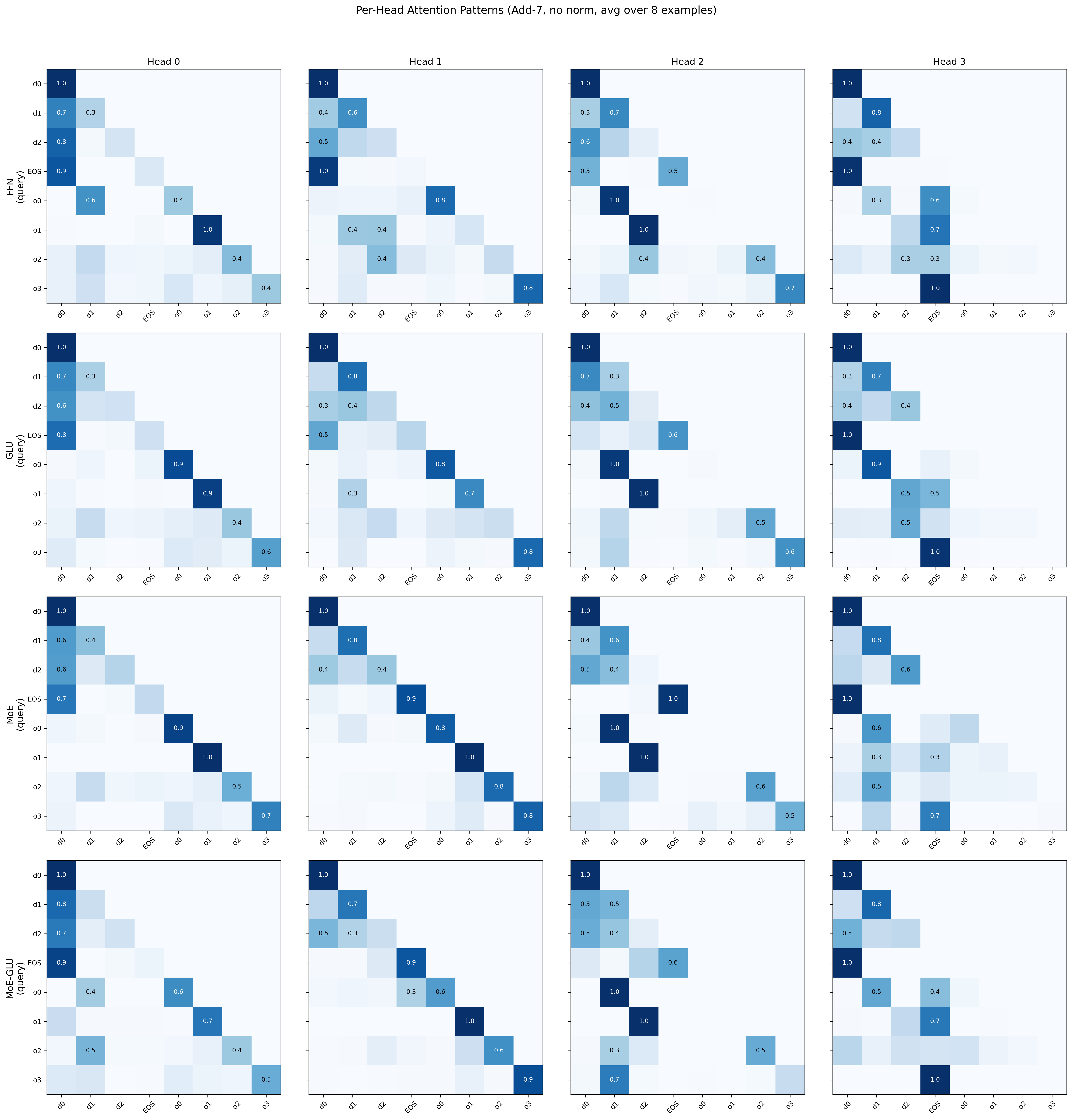}
  \caption{Attention pattern heatmaps across all 4 variants on add-7 (no norm). MoE develops cleaner head specialization.}
  \label{fig:attention-patterns}
\end{figure}

\begin{figure}[htbp]
  \centering
  \includegraphics[width=0.85\linewidth]{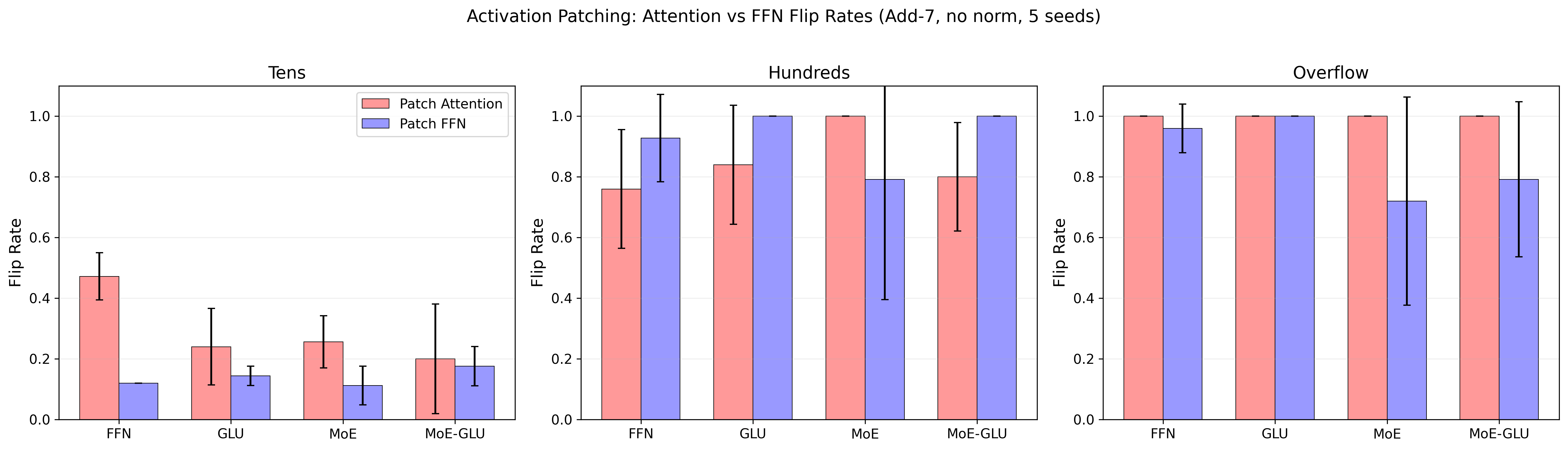}
  \caption{Activation patching flip rates by position and variant on add-7. In most position--variant cells, patching attention flips predictions more often than patching FFN. The hundreds position is a notable exception: FFN patching produces a higher flip rate for FFN, GLU, and MoE-GLU. Error bars: $\pm 1$ std across 5 seeds.}
  \label{fig:activation-patching}
\end{figure}

%% ---- D: Grokking and Optimization-Side Effects ----
\section{Grokking and Optimization-Side Effects}
\label{app:grokking-controls}

MoE accelerates grokking on modular addition, but a dropout baseline produces a comparable speedup, so we report this as an optimization-side effect of sparse training rather than a primary contribution. This appendix reports the data, the controls, and the alternative-explanation analysis.

\subsection{MoE Grokking Speedup on Modular Addition}
\label{app:grokking-speedup}

\begin{figure}[htbp]
  \centering
  \begin{subfigure}[t]{0.48\linewidth}
    \centering
    \includegraphics[width=\linewidth]{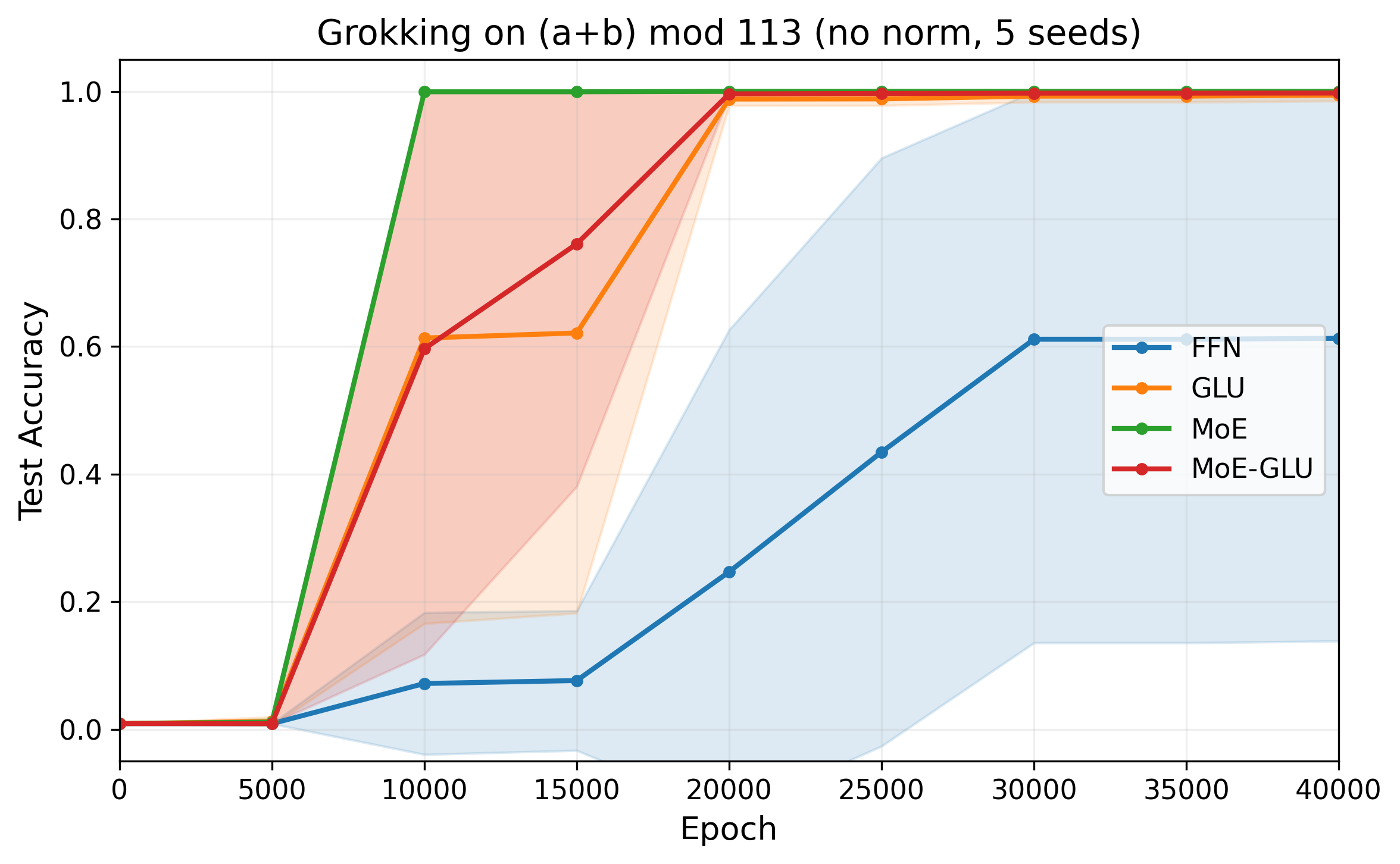}
    \caption{Grokking timelines (5 seeds per variant shown). MoE reaches 99\% test accuracy at ${\sim}8{,}700$ epochs vs.\ ${\sim}24{,}000$ for dense. Extended to $n{=}20$: MoE 20/20 grok vs.\ FFN 12/20 within 40k epochs.}
    \label{fig:grokking}
  \end{subfigure}
  \hfill
  \begin{subfigure}[t]{0.48\linewidth}
    \centering
    \includegraphics[width=\linewidth]{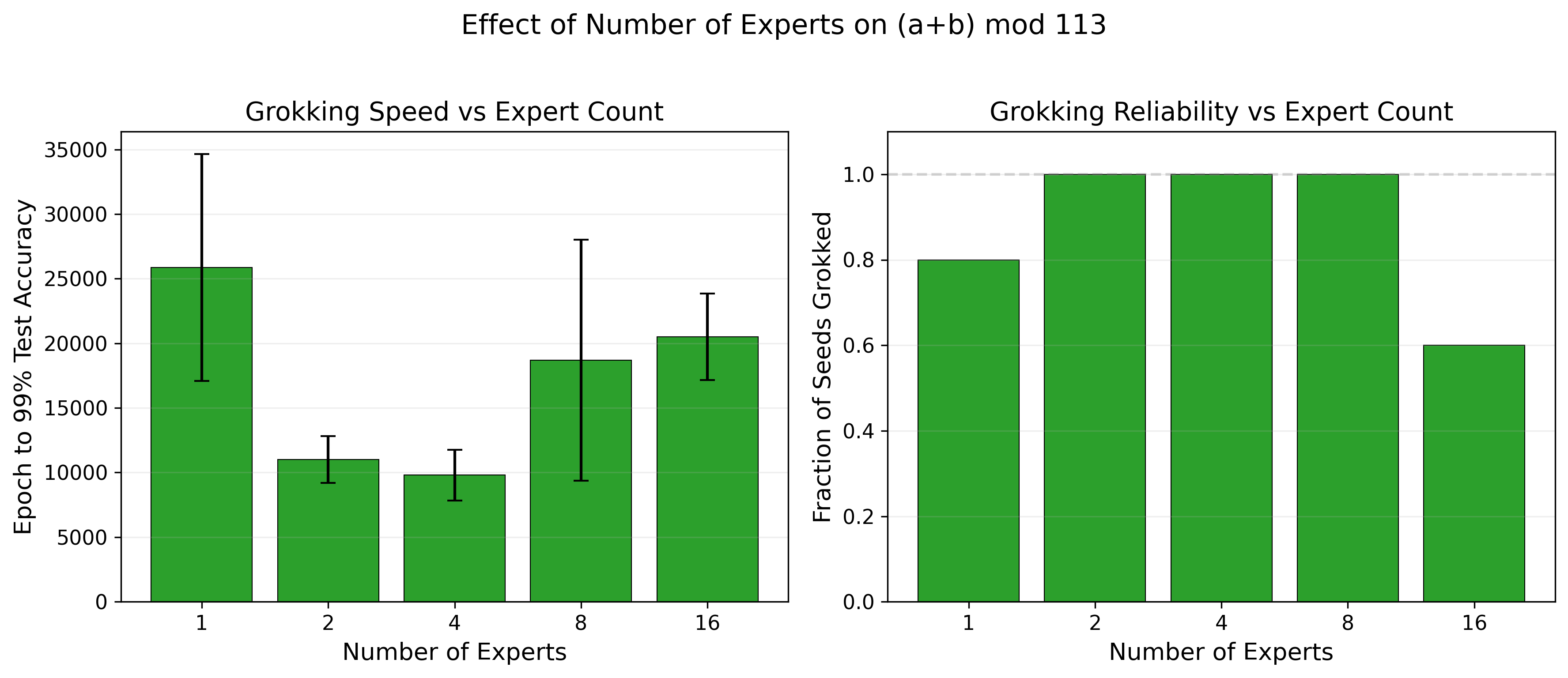}
    \caption{Number of experts. $E{=}1$ (auxiliary loss, no routing) does not accelerate grokking. The effect requires routing to multiple experts.}
    \label{fig:num-experts}
  \end{subfigure}
  \caption{\textbf{MoE accelerates grokking on modular addition} (5 seeds). \textbf{(a)}~Shaded regions show $\pm 1$ standard deviation across seeds. \textbf{(b)}~The $E{=}1$ control has the auxiliary loss but no routing; it does not accelerate grokking, confirming sparse routing itself contributes. Dropout${}=0.3$ achieves a faster speedup than MoE (App.~\ref{app:reg-baselines}). All error bars and shaded regions in this paper show $\pm 1$ std across 5 seeds unless otherwise noted.}
  \label{fig:grokking-controls}
\end{figure}

\paragraph{Reliability and speedup.}
With $n{=}20$ seeds per variant, MoE reaches $99\%$ test accuracy in all $20/20$ runs within 40k epochs, while FFN groks in only $12/20$ (the remaining $8$ stay below $99\%$ at the cutoff). Treating the $8$ censored FFN seeds as tied at 40k gives Mann-Whitney $U{=}390$, $p{=}1.3\times 10^{-7}$ (one-sided). Among seeds that do grok, MoE's median epoch-to-$99\%$ is $8{,}400$ vs.\ $26{,}300$ for FFN, a $3.12\times$ speedup. The reliability gap ($100\%$ vs.\ $60\%$) is the more robust claim because it does not depend on the cutoff choice; the dense-FFN failure distribution is bimodal (either grok or don't), not a slow tail. The original 5-seed timelines (Fig.~\ref{fig:grokking}) replicate at the larger $n$.

\paragraph{Disentangling the mechanism.}
The $E{=}1$ control has the auxiliary loss but no routing decision; it does not accelerate grokking (Fig.~\ref{fig:num-experts}), confirming that routing to multiple experts is necessary. Top-2 routing substantially reduces the advantage entirely (0/5 seeds; App.~\ref{app:topk}): when every token accesses two experts, the bottleneck disappears. Hard top-1 routing thus appears to have a regularizing effect, consistent with \citet{zhang2022moefication}'s observation that FFN layers exhibit natural sparsity that MoE makes explicit. The advantage grows with model width ($1.1\times$ at $d{=}64$, $2.7\times$ at $d{=}256$; App.~\ref{app:width-scaling}). However, dropout achieves a faster speedup than MoE on this task (App.~\ref{app:reg-baselines}), so we cannot claim that MoE's routing produces a uniquely valuable regularization signal, only that it produces some regularization, which the $E{=}1$ and top-2 controls show specifically requires hard sparse routing across multiple experts.

\subsection{Random Routing on Modular Addition}                                    
\label{app:modadd-grokking}                                                   
We compare learned-routing MoE against frozen random-routing MoE on modular addition (5 seeds each, $d_m{=}128$, $E{=}4$, top-1, 40k-epoch budget). Learned routing groks in all $5/5$ seeds, with epoch-to-99\% between 5k and 10k for every seed (median ${\sim}10{,}000$). Random routing groks in $4/5$ seeds, with epoch-to-99\%  spanning ${\sim}25$k to ${\sim}40$k (median ${\sim}27{,}500$, mean ${\sim}30{,}000$).  The fifth seed remains at chance ($0.9\%$ test accuracy) at the 40k-epoch cutoff. Random routing therefore preserves the eventual solution in most seeds but loses both the speed and the reliability of a learned router. This contrasts with the redistribution measure (most notably no-FFN accuracy on add-7), where learned and random routing are statistically indistinguishable (Sec.~\ref{sec:routing-causal}, Tab.~\ref{tab:random-routing}).  Learned routing affects when (and whether) grokking happens, but not how the trained network divides labor between attention and FFN.

\FloatBarrier
\subsection{Regularization Baselines}
\label{app:reg-baselines}

To test whether MoE's grokking advantage stems from a regularization effect, we compare against dropout and weight decay baselines on modular addition.

\begin{figure}[htbp]
  \centering
  \includegraphics[width=0.85\linewidth]{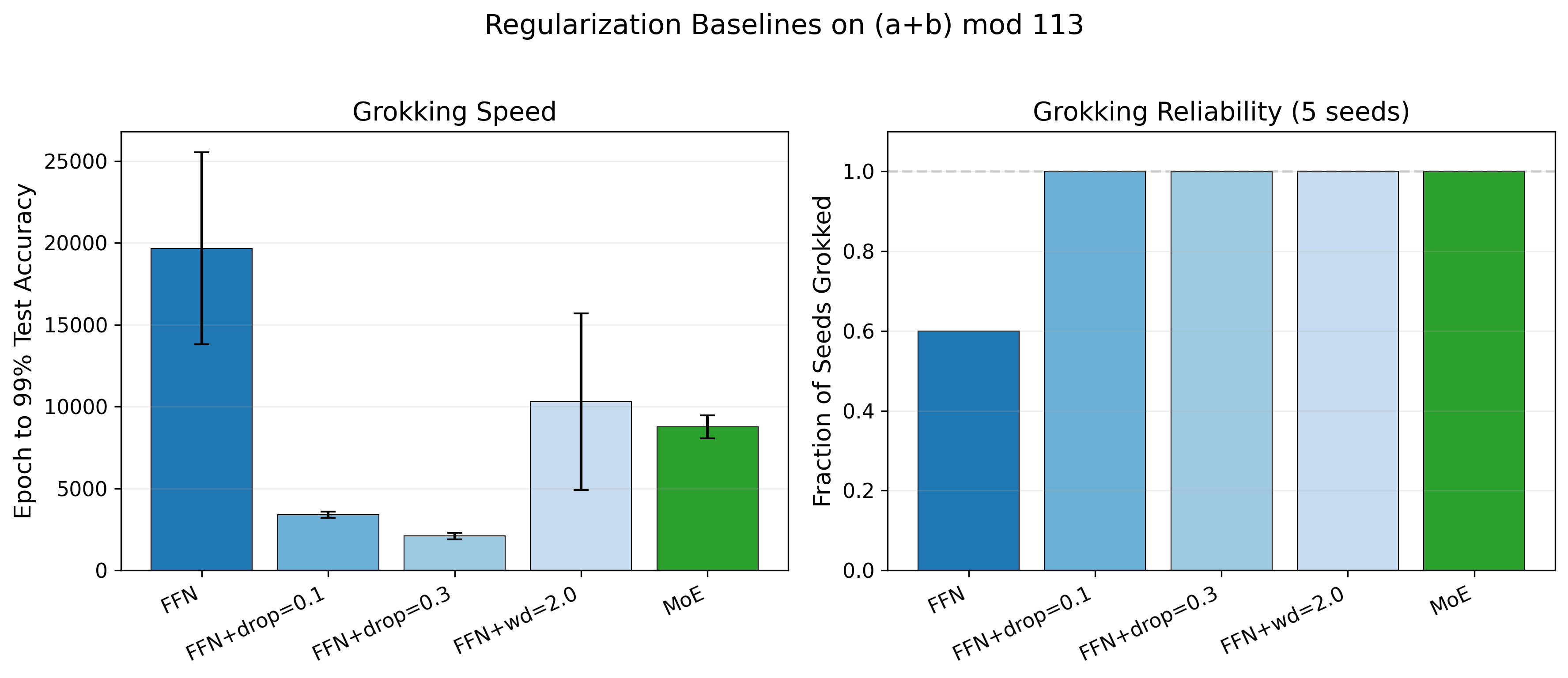}
  \caption{Regularization baselines on modular addition. Dropout${}=0.3$ achieves median grokking at ${\sim}2{,}100$ epochs, faster than MoE, confirming a regularization component. Error bars: $\pm 1$ std across 5 seeds.}
  \label{fig:reg-baselines}
\end{figure}

\FloatBarrier
\subsection{Top-$k$ Routing}
\label{app:topk}

Top-2 routing gives each token access to two experts, alleviating the per-token capacity bottleneck. The top-2 control uses the same architecture (same $E$, same $h_E$) with $k{=}2$ routing, isolating the routing-$k$ axis at fixed architecture. If the grokking advantage requires hard routing decisions and tight per-token capacity, top-2 should eliminate it.

\begin{figure}[htbp]
  \centering
  \includegraphics[width=0.85\linewidth]{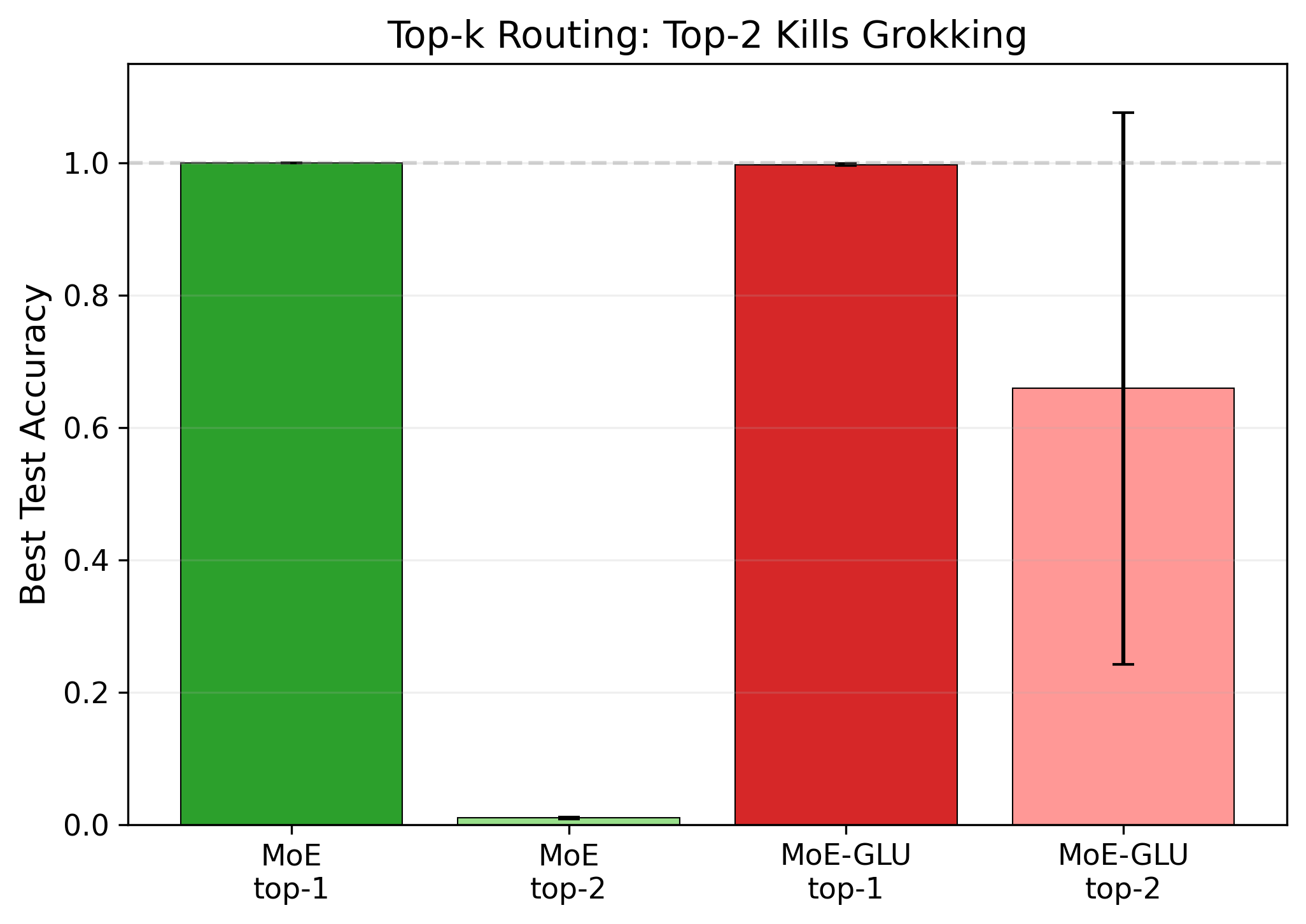}
  \caption{Top-1 vs.\ top-2 routing on modular addition. Top-2 eliminates the grokking advantage (0/5 seeds within 40k epochs). Hard routing is essential. Error bars: $\pm 1$ std across 5 seeds.}
  \label{fig:topk}
\end{figure}

\FloatBarrier
\subsection{Width Scaling}
\label{app:width-scaling}

We test whether MoE's grokking advantage scales with model width by training at $d \in \{64, 128, 256\}$.

\begin{figure}[htbp]
  \centering
  \includegraphics[width=0.85\linewidth]{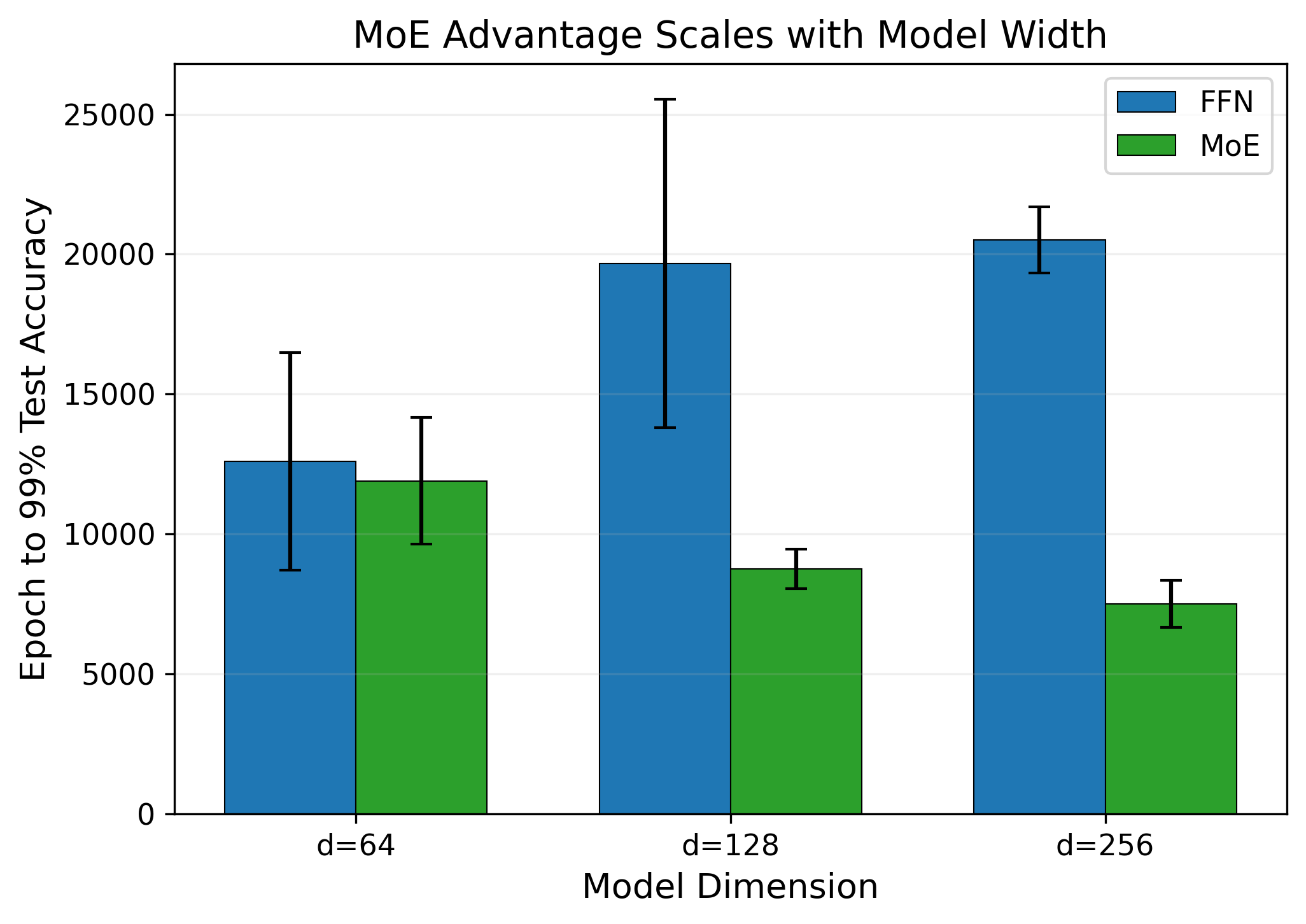}
  \caption{Width scaling of MoE grokking advantage on modular addition. Mean epoch-to-$99\%$ across 5 seeds (FFN at $d{=}128$ averages over the $3/5$ seeds that grok within the 40k cutoff): FFN $12{,}600$ vs.\ MoE $11{,}900$ at $d{=}64$ ($1.1\times$), FFN $19{,}667$ vs.\ MoE $8{,}760$ at $d{=}128$ ($2.2\times$), FFN $20{,}500$ vs.\ MoE $7{,}500$ at $d{=}256$ ($2.7\times$). Error bars: $\pm 1$ std across 5 seeds.}
  \label{fig:width-scaling}
\end{figure}

%% ---- E: Expert Specialization ----
\section{Expert Specialization by Operation Type}
\label{app:specialization}

We treat expert-operation specialization as a secondary observation rather than a primary contribution: the NMI mean is comparable to its standard deviation, the effect appears only on add-7 (not modadd), and exceeding the random-routing chance NMI of $0.0008$ is a low bar. We report the data here for completeness.

\begin{figure}[htbp]
  \centering
  \includegraphics[width=0.85\linewidth]{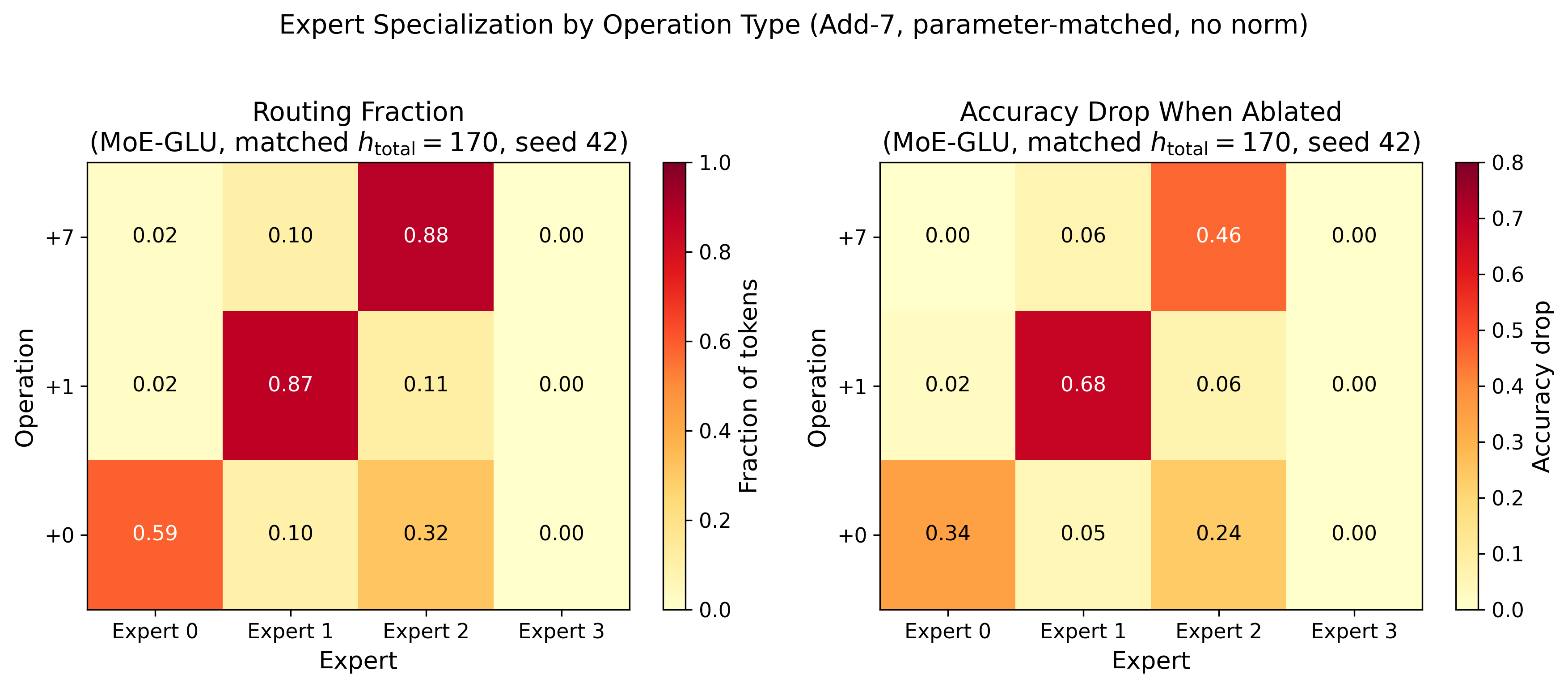}
  \caption{\textbf{Expert-operation routing on add-7} (MoE-GLU, seed 42, the strongest-specialization seed; see App.~\ref{app:perseed-routing} for all 5 seeds). \textbf{Left}: routing heatmap showing expert~2 receives $88\%$ of $+7$ tokens. \textbf{Right}: expert ablation confirms causal impact. Specialization varies substantially across seeds (NMI $= 0.28 \pm 0.20$).}
  \label{fig:expert-routing}
\end{figure}

On add-7, MoE-GLU achieves normalized mutual information of $0.28 \pm 0.20$ (mean $\pm$ std across 5 seeds) between expert assignment and operation type, with MoE at $0.26 \pm 0.18$. Under random routing, chance NMI is $0.0008 \pm 0.0005$ (1000 simulations), so even the weakest observed seed exceeds chance by an order of magnitude, but the variance across seeds is nearly equal to the mean: in some seeds one expert captures nearly all $+7$ tokens, while in others the routing is more diffuse (App.~\ref{app:perseed-routing}). We show a clear case (MoE-GLU, seed 42) in Fig.~\ref{fig:expert-routing}, where expert~2 receives $88\%$ of $+7$ tokens and expert~1 receives $87\%$ of $+1$ tokens. Expert ablation confirms causal functional assignments in this seed; aggregated across all 5 seeds, ablating the $+7$-dominant expert drops $+7$ accuracy by $0.38 \pm 0.12$ while barely affecting $+1$ ($0.06 \pm 0.07$) for MoE-GLU, and $0.31 \pm 0.17$ / $0.07 \pm 0.06$ for MoE.

We read this as a tendency modulated by initialization, not a robust property of MoE. The high variance likely reflects early routing decisions: whichever expert receives more $+7$ tokens initially gets reinforced into that role, and small initialization differences cascade into different patterns. This is a mechanistic micro-version of the specialization observed at scale by \citet{dai2024deepseekmoe}, where fine-grained routing produces experts with distinct roles; in our controlled setting we can verify causality via ablation. On modular addition, routing shows no correlation with input structure. Specialization emerges only on tasks with discrete operation categories, not continuous algebraic structure.

\FloatBarrier
\subsection{Per-Seed Expert Routing Variability}
\label{app:perseed-routing}

\begin{figure}[htbp]
  \centering
  \includegraphics[width=0.85\linewidth]{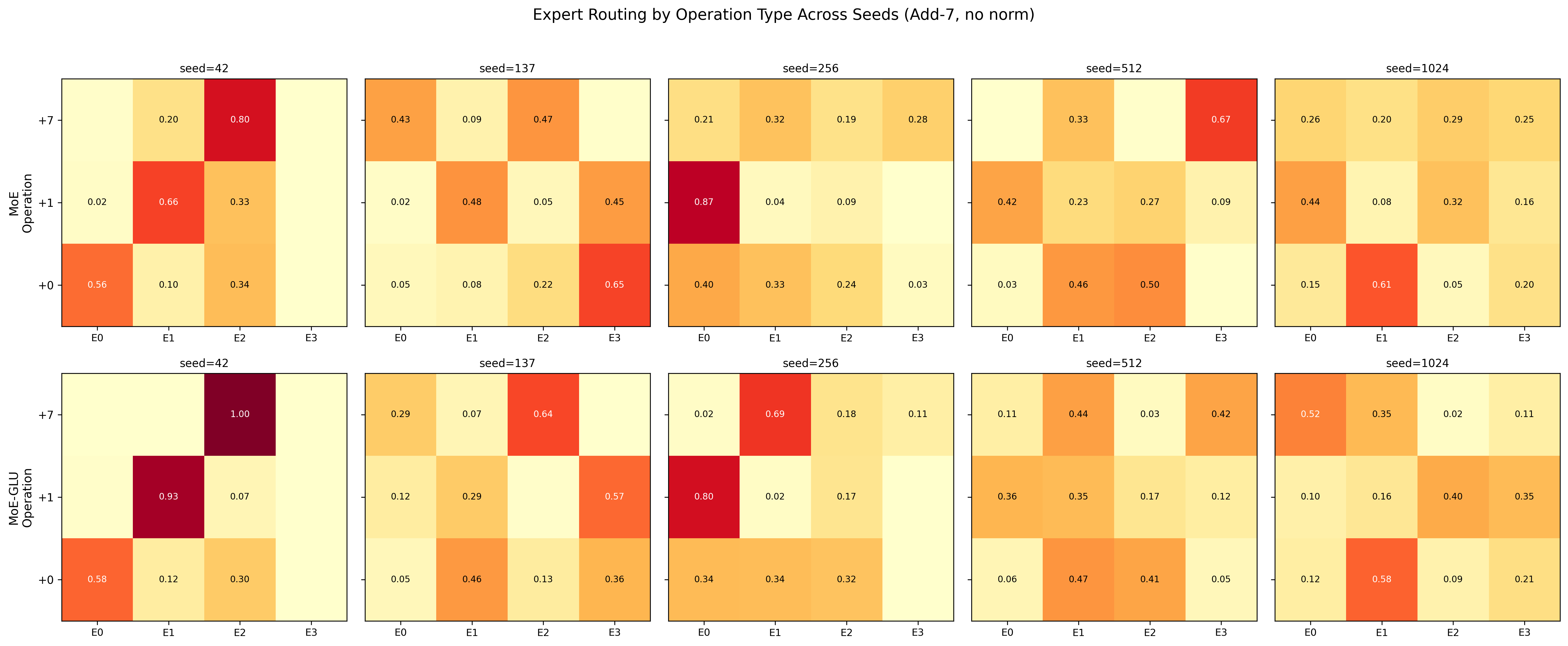}
  \caption{Expert routing heatmaps across all 5 seeds for MoE and MoE-GLU on add-7. Specialization direction is consistent (one expert gravitates toward $+7$) but degree varies substantially across seeds.}
  \label{fig:perseed-routing}
\end{figure}

The aggregate NMI of $0.28 \pm 0.20$ reported above hides substantial seed-to-seed variability. Fig.~\ref{fig:perseed-routing} shows the full per-seed heatmaps for both MoE and MoE-GLU on add-7.

%% ---- F: GLU Analysis ----
\section{GLU Analysis Details}
\label{app:glu-details}

\subsection{Gate Decomposition}
\label{app:glu-decomposition}

\begin{figure}[htbp]
  \centering
  \includegraphics[width=0.85\linewidth]{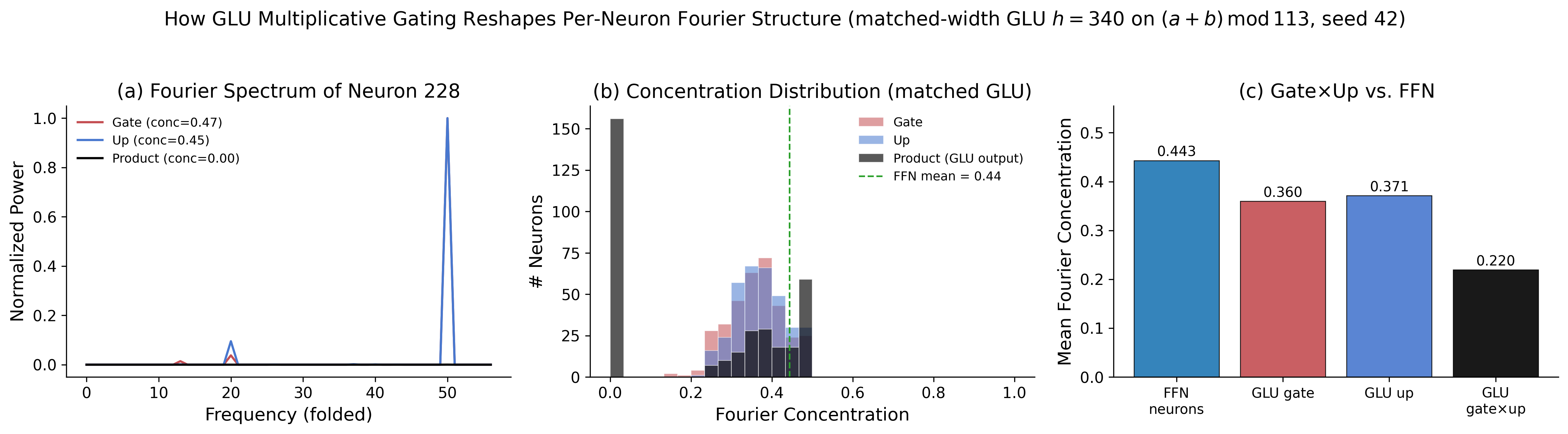}
  \caption{GLU decomposition on modular addition (matched-width $h{=}340$, seed 42). Mean per-neuron Fourier concentration: dense FFN $0.443$, GLU gate $0.360$, GLU up-projection $0.371$, GLU product (gate $\odot$ up) $0.220$. Gate and up-projection individually retain FFN-like per-neuron structure; the multiplicative interaction is what halves per-neuron concentration. Linear probes achieve 100\% from all components, so task information is preserved across the gate. This identifies the multiplicative gate, not low pre-multiplication structure, as the operation that rotates Fourier content out of the neuron basis.}
  \label{fig:glu-decomposition}
\end{figure}

\FloatBarrier
\subsection{Subspace Analysis (PCA)}
\label{app:glu-pca}

Activations are mean-centered before computing SVD. We project onto the top 10 principal components and measure each component's Fourier concentration by grouping projected values by $(a{+}b) \bmod p$, averaging within each group, and computing the fraction of spectral power at the dominant FFT frequency.

The per-PC breakdown reveals that the high concentration is not isolated to a single principal component but broadly distributed. At parameter matched widths, GLU's top-6 PCs all exceed $0.36$ (PCs 0, 1, 3, 4, 5 exceed $0.43$), with the top-3 PCs explaining $96.5\%$ of variance (vs.\ $85.8\%$ for FFN's top-10). GLU is notably low-rank: top-10 PCs capture $100.0\%$ of variance. This means the Fourier structure occupies a low-dimensional subspace that is simply rotated relative to the neuron basis.

\paragraph{A note on per-neuron vs.\ top-PC concentration.} The two metrics measure complementary kinds of Fourier structure. Per-neuron concentration is high when individual neurons each respond to a specific frequency (as \citet{nanda2023progress} found in FFN); top-PC concentration is high when the dominant variance directions align with single frequencies, regardless of whether individual neurons are frequency-specific. For matched-width FFN, both metrics are moderately high ($0.34$ per-neuron, $0.41$ top-1 PC), reflecting partial per-neuron tuning whose specificity is diluted in the principal directions because PCA averages across neurons responding to different frequencies. For GLU, no such per-neuron tuning exists ($0.17$ per-neuron); the gate scrambles each neuron's response, but the dominant variance directions still align tightly with frequencies ($0.49$ top-1 PC). PCA recovers the structure that the per-neuron view loses, putting GLU and MoE-GLU in a regime where Fourier circuits are rotated into a low-dimensional subspace rather than embedded in the neuron basis.

\begin{table}[htbp]
\centering
\small
\begin{tabular}{lcccc}
\toprule
& \textbf{FFN} & \textbf{GLU} & \textbf{MoE} & \textbf{MoE-GLU} \\
\midrule
Per-neuron concentration & 0.34 & 0.17 & 0.48 & 0.16 \\
Top-1 PC concentration & 0.41 & \textbf{0.49} & 0.49 & 0.46 \\
Mean top-6 PC concentration & 0.36 & \textbf{0.46} & 0.46 & 0.44 \\
Top-10 expl.\ variance & 85.8\% & 100.0\% & 97.7\% & 98.2\% \\
\bottomrule
\end{tabular}
\caption{Fourier concentration breakdown at total-parameter-matched widths (5 seeds, modadd). Per-neuron concentration measures individual neuron frequency selectivity. Top-1 PC and mean top-6 PC concentrations measure subspace-level structure. GLU's per-neuron concentration is roughly half of FFN's, but its principal components match or exceed the others' (0.49 top-1, 0.46 mean top-6), confirming structure is preserved in a low-dimensional subspace. GLU is also the most low-rank (100.0\% variance in top-10 PCs vs.\ 85.8\% for FFN). Per-neuron values agree with Tab.~\ref{tab:fourier-neuron-control}.}
\label{tab:glu-pca}
\end{table}

\FloatBarrier
\subsection{Width and Parameter Controls for GLU}
\label{app:glu-width-controls}

We run two controls to rule out that GLU's low per-neuron concentration is just an artifact of width or parameter count. First, we train a dense FFN at half width (128 instead of 256 hidden units) to match the original GLU's neuron count. Second, on modular addition, we train a GLU with intermediate dim 340, which matches dense FFN's parameter count exactly (130{,}560 vs.\ 131{,}712).

\begin{table}[htbp]
\centering
\small

\setlength{\tabcolsep}{4pt}

\renewcommand{\arraystretch}{1.08}

\begin{tabularx}{\linewidth}{Xccc}

\toprule

\textbf{Variant} & \textbf{Hidden} & \textbf{FFN params} & \textbf{Per-neuron conc.} \\

\midrule

\multicolumn{4}{l}{\textit{add-7 ($d_m{=}64$; default-width controls; Dense FFN is single-seed, others 5 seeds)}} \\

\hspace{1em}Dense FFN              & 256 & 33{,}088  & $0.44$ \\

\hspace{1em}Dense FFN (narrow)     & 128 & 16{,}640  & $0.45 \pm 0.01$ \\

\hspace{1em}GLU \textit{(default, undersized)} & 128 & 24{,}576 \textit{($0.74\times$)} & $0.07 \pm 0.04$ \\

\midrule

\multicolumn{4}{l}{\textit{modular addition ($d_m{=}128$; total-parameter-matched; 5 seeds)}} \\

\hspace{1em}Dense FFN              & 512 & 131{,}712 & $0.34 \pm 0.13$ \\

\hspace{1em}MoE ($E{=}4$)          & $h_E{=}128$ & 132{,}608 & $0.48 \pm 0.01$ \\

\hspace{1em}\textbf{GLU param-matched}     & 340 & 130{,}560 & $\mathbf{0.17 \pm 0.12}$ \\

\hspace{1em}\textbf{MoE-GLU param-matched} & $h_E{=}85$ & 131{,}072 & $\mathbf{0.16 \pm 0.10}$ \\

\bottomrule

\end{tabularx}

% \begin{tabular}{lccc}
% \toprule
% \textbf{Variant} & \textbf{Hidden} & \textbf{FFN params} & \textbf{Per-neuron conc.} \\
% \midrule
% \textit{add-7 ($d_m{=}64$, default-width controls; Dense FFN row is single-seed, others are 5 seeds)} & & & \\
% \hspace{1em}Dense FFN              & 256 & 33{,}088  & $0.44$ \\
% \hspace{1em}Dense FFN (narrow)     & 128 & 16{,}640  & $0.45 \pm 0.01$ \\
% \hspace{1em}GLU \textit{(default, undersized)} & 128 & 24{,}576 \textit{($0.74\times$ dense)} & $0.07 \pm 0.04$ \\
% \midrule
% \textit{modular addition ($d_m{=}128$, total-parameter-matched, 5 seeds)} & & & \\
% \hspace{1em}Dense FFN              & 512 & 131{,}712 & $0.34 \pm 0.13$ \\
% \hspace{1em}MoE ($E{=}4$)          & $h_E{=}128$ & 132{,}608 & $0.48 \pm 0.01$ \\
% \hspace{1em}\textbf{GLU param-matched}     & 340 & 130{,}560 & $\mathbf{0.17 \pm 0.12}$ \\
% \hspace{1em}\textbf{MoE-GLU param-matched} & $h_E{=}85$ & 131{,}072 & $\mathbf{0.16 \pm 0.10}$ \\
% \bottomrule
% \end{tabular}
\caption{Per-neuron Fourier concentration controls. \textbf{Add-7 (top):} halving the dense FFN's hidden dimension to match the default-width GLU's 128 neurons leaves per-neuron concentration unchanged ($0.45$ vs.\ $0.44$); the order-of-magnitude drop for default-width GLU ($0.07$) is caused by the multiplicative gate, not neuron count. Note this default-width GLU is at $0.74\times$ dense FFN parameters (undersized), so the matched-width modadd rows in this table are the apples-to-apples comparison. \textbf{Modular addition (bottom):} at total-parameter-matched widths and aggregated across 5 seeds, the gate-and-no-gate variants separate clearly: FFN/MoE concentrate around $0.34$/$0.48$, while GLU/MoE-GLU collapse to $0.17$/$0.16$ at the same parameter budget. The FFN seed-mean std of $0.13$ reflects one outlier seed (concentration $0.09$); FFN over the other four seeds means $\approx 0.40$. Means $\pm$ seed-mean std, $n{=}5$ seeds.}
\label{tab:fourier-neuron-control}
\end{table}

\paragraph{Width-scan control for MoE-GLU.}
The main results (Fig.~\ref{fig:component-ablation}) use MoE-GLU at $h_E{=}42$ on add-7 and $h_E{=}85$ on modadd / histogram (total expert width matches GLU's $\tfrac{2}{3}h_{\text{dense}}$). To confirm the redistribution is not specific to this narrow per-expert width, we also evaluate the older default of $h_E{=}h_{\text{dense}}/E$ ($h_E{=}64$ on add-7, $h_E{=}128$ on modadd; this default makes total MoE-GLU params $\approx 1.5\times$ dense). On add-7, default-width MoE-GLU retains $32.5 \pm 6.0\%$ no-FFN accuracy ($n{=}5$) vs.\ $41.7 \pm 2.8\%$ at the total-matched $h_E{=}42$, both substantially above dense FFN's $11.9\%$. On modadd, default-width retains $2.9 \pm 1.4\%$ no-FFN vs.\ $2.6 \pm 1.3\%$ at total-matched $h_E{=}85$ (Welch $p{>}0.7$). A wider per-active-matched run ($h_E{=}170$ on add-7, $h_E{=}340$ on modadd, giving $4\times$ dense total params) is reported as a separate control in App.~\ref{app:per-active-controls}.

%% ---- G: Capacity Controls ----
\section{Capacity and Generalization Controls}
\label{app:capacity}

\subsection{Frozen Component Training}
\label{app:frozen}

To distinguish training-time redistribution from capacity limits, we train each variant with one component frozen at random initialization.

\begin{table}[htbp]
\centering
\small
\begin{tabular}{lcccccc}
\toprule
& \multicolumn{3}{c}{\textbf{Frozen Attention}} & \multicolumn{3}{c}{\textbf{Frozen FFN}} \\
\cmidrule(lr){2-4} \cmidrule(lr){5-7}
\textbf{Variant} & Add-7 & ModAdd & Hist. & Add-7 & ModAdd & Hist. \\
\midrule
FFN     & 100\% & 100\%  & $98.3 \pm 0.6$\% & 100\% & $3.9 \pm 3.8$\% & $96.6 \pm 3.4$\% \\
GLU     & 100\% & 100\%  & $98.9 \pm 0.5$\% & 100\% & $58.5 \pm 46.4$\% & $95.6 \pm 0.7$\% \\
MoE     & 100\% & 99.8\% & $99.3 \pm 0.1$\% & 100\% & $1.6 \pm 1.2$\% & $96.0 \pm 2.4$\% \\
MoE-GLU & 100\% & $83.5 \pm 23.4$\% & $99.4 \pm 0.1$\% & 100\% & $40.5 \pm 44.7$\% & $94.1 \pm 3.1$\% \\
\bottomrule
\end{tabular}
\caption{Frozen component training (5 seeds). Each variant is trained with either attention or FFN frozen at random initialization. FFN alone solves add-7 and nearly solves histogram ($96$-$99\%$). Attention alone solves add-7 but largely fails modular addition for dense FFN and MoE variants ($<4\%$). Histogram is solvable by either component alone ($>94\%$ in all conditions), consistent with the minimal redistribution observed for this task. GLU variants show high variance under frozen FFN on modular addition ($58.5 \pm 46.4\%$, $40.5 \pm 44.7\%$), suggesting GLU's random bilinear projection occasionally provides a basis attention can exploit.}
\label{tab:frozen}
\end{table}

\FloatBarrier
\subsection{Held-Out Generalization}
\label{app:generalization}

\begin{table}[htbp]
\centering
\small
\begin{tabular}{lcccc}
\toprule
\textbf{Experiment} & FFN & GLU & MoE & MoE-GLU \\
\midrule
Excl.\ $L \geq 2$ (train) & 100\% & 100\% & 100\% & 100\% \\
Excl.\ $L \geq 2$ (held-out) & 0\% & 0\% & 0\% & 0\% \\
\midrule
Excl.\ ones $\geq 3$ (train) & 100\% & 100\% & 100\% & 100\% \\
Excl.\ ones $\geq 3$ (held-out) & 0\% & 0\% & 0\% & 0\% \\
\midrule
Excl.\ tens${=}9$ (train) & 100\% & 100\% & 100\% & 100\% \\
Excl.\ tens${=}9$ (held-out) & 24.0\% & 2.0\% & \textbf{26.4\%} & 16.0\% \\
\bottomrule
\end{tabular}
\caption{Held-out generalization on add-7 (5 seeds). No variant generalizes unseen operations (Exps 1-2). For positional transfer (Exp 3), MoE generalizes best ($26.4 \pm 4.5$\%) while GLU fails ($2.0 \pm 4.0$\%).}
\label{tab:generalization}
\end{table}

\FloatBarrier
\subsection{Narrow Dense FFN Control (Modular Addition)}
\label{app:narrow-modadd}

To test whether MoE's optimization advantage on modular addition stems from reduced per-token capacity or from routing, we train a dense FFN with intermediate dimension 128 (matching MoE's per-expert width of $512/4$).

\begin{figure}[htbp]
  \centering
  \includegraphics[width=0.85\linewidth]{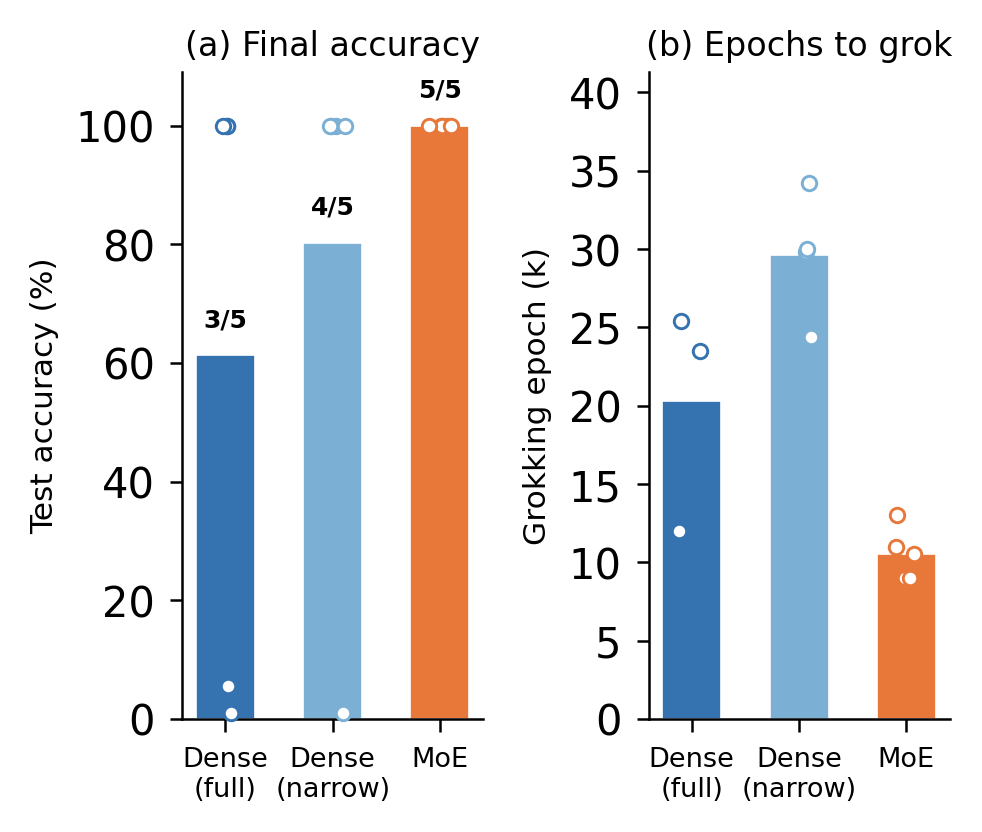}
  \caption{Narrow dense FFN vs.\ full-width dense FFN vs.\ MoE on modular addition (5 seeds). \textbf{(a)}~The narrow FFN solves the task in 4/5 seeds (vs.\ 3/5 for full-width), showing that reduced capacity alone does not prevent grokking. \textbf{(b)}~Among solved seeds, MoE groks fastest (${\sim}10$k epochs, 5/5 seeds), while the narrow FFN groks substantially later (${\sim}30$k epochs) and later than the full-width FFN; the MoE and narrow-FFN solved-seed epoch ranges do not overlap in this run. \emph{Reducing per-token capacity alone (narrow FFN) slows grokking relative to full-width FFN, while sparse partitioning at the same per-token capacity (MoE) accelerates it. The grokking speedup therefore points to sparse partitioning, not reduced per-token capacity alone.}}
  \label{fig:narrow-modadd}
\end{figure}

\FloatBarrier
\subsection{Parameter Matching vs.\ FLOP Matching}
\label{app:param-vs-flop}

The scaling-MoE literature typically FLOP-matches rather than total-parameter-matches \citep{fedus2022switch, jiang2024mixtral} because its research question is whether sparsity can recover dense-level quality at lower active compute; there, the per-token capacity reduction is the cost to be minimized. Our research question inverts this: the per-token capacity reduction is the mechanism under study. A FLOP-matched MoE would set each expert's hidden width equal to the dense FFN's, removing the sparsity bottleneck by construction and eliminating the effect we measure. We therefore total-parameter-match plain MoE to dense FFN. Our random routing control (Sec.~\ref{sec:routing-causal}) further rules out the alternative that the effect requires learned adaptation to a tight FLOP budget: routing weights are frozen at initialization, yet the redistribution pattern is statistically indistinguishable from learned routing. The effect is driven by sparse partitioning per se, not by learned compensation for reduced active compute.

For MoE-GLU we report total-parameter-matched-as-GLU variants throughout the main results (total expert width $\tfrac{2}{3}h_{\text{dense}}$, per-expert $h_E = \tfrac{2}{3}h_{\text{dense}}/E$, per-active capacity $1/E$ of dense FFN). This is the strictest available matching: MoE-GLU enters the redistribution comparison with both the same total parameter budget and strictly less per-active capacity than dense. A per-active-matched control ($h_E{=}\tfrac{2}{3}h_{\text{dense}}$ per expert, $\approx 4\times$ dense total params) is reported in App.~\ref{app:per-active-controls} (Tab.~\ref{tab:per-active-controls}): the redistribution direction holds at FLOP parity but is roughly halved on add-7, providing an upper bound on the contribution of sparse partitioning when per-token capacity is held at dense level.

\FloatBarrier
\subsection{Per-active matched controls (FLOP parity)}
\label{app:per-active-controls}

We re-train MoE-GLU at the alternative matching convention used by scaling work \citep{fedus2022switch, jiang2024mixtral}: each expert's hidden width $h_E$ matches dense FFN's per-token FLOPs, giving FLOP parity per token at the cost of $\approx 4\times$ dense total parameters. Concretely, we set the codebase's total \texttt{intermediate\_dim} to $E \cdot h_{\text{glu}}$, so the per-expert width is $h_E{=}170$ on add-7 (intermediate\_dim$=680$) and $h_E{=}340$ on modadd / histogram (intermediate\_dim$=1360$). All other architecture and optimization settings match the headline runs.

\begin{table}[h]
\centering
\small
\begin{tabular}{llrrrr}
\toprule
\textbf{Task} & \textbf{Convention} & \textbf{$h_E$} & \textbf{Total FFN params} & \textbf{Normal} & \textbf{No-FFN} \\
\midrule
add-7              & total-param matched (headline) & $42$  & $\phantom{0}32{,}512$ ($1.0\times$) & $100.0\%$ & $41.7 \pm 2.8\%$ \\
add-7              & per-active matched (FLOP)      & $170$ & $130{,}816$ ($4.0\times$)           & $100.0\%$ & $\mathbf{18.3 \pm 5.6\%}$ \\
\midrule
modular addition   & total-param matched (headline) & $85$  & $131{,}072$ ($1.0\times$)           & $99.4\%$  & $\phantom{0}2.6 \pm 1.3\%$ \\
modular addition   & per-active matched (FLOP)      & $340$ & $522{,}752$ ($4.0\times$)           & $99.7\%$  & $\phantom{0}\mathbf{1.6 \pm 0.5\%}$ \\
\midrule
histogram          & total-param matched (headline) & $85$  & $131{,}072$ ($1.0\times$)           & $99.8\%$  & $10.4 \pm 0.2\%$ \\
histogram          & per-active matched (FLOP)      & $340$ & $522{,}752$ ($4.0\times$)           & $99.9\%$  & $\mathbf{10.2 \pm 0.2\%}$ \\
\bottomrule
\end{tabular}
\caption{MoE-GLU at total-parameter matching (paper headline, same total params as dense, $1/E$ per-active capacity) vs.\ per-active matching (FLOP parity per token, $4\times$ dense total params). 5 seeds each; values are mean $\pm$ standard deviation.}
\label{tab:per-active-controls}
\end{table}

\paragraph{Interpretation.} The MoE-GLU redistribution direction holds at both matching conventions but its magnitude is convention-dependent on add-7. Three observations:
\begin{enumerate}
\item \textbf{On add-7, per-active matching more than halves the redistribution} ($41.7\% \to 18.3\%$). Increasing per-token capacity from $1/E$ of dense to full dense, while keeping the sparse-partitioning architecture, removes most of the no-FFN gap. This is consistent with the narrow-dense-FFN control in the main text (Sec.~\ref{sec:routing-causal}), which attributes a substantial fraction of the dense-to-MoE-GLU gap to per-token capacity reduction. The per-active-matched MoE-GLU still retains $18.3\%$ no-FFN, above dense FFN's $11.9\%$, so sparse partitioning continues to drive a residual redistribution effect even at FLOP parity.
\item \textbf{On modular addition, both conventions give negligible redistribution} ($2.6\%$ vs.\ $1.6\%$, both close to dense FFN's $1.3\%$). Modular addition's reliance on Fourier circuits exceeds attention's expressive capacity \citep{nanda2023progress}, so attention cannot absorb FFN computation regardless of bottleneck severity, and the matching convention has little effect.
\item \textbf{On histogram, both conventions give identical no-FFN} ($10.4\%$ vs.\ $10.2\%$, both at the chance line for $L{=}10$ tokens). Histogram is the non-substitutable control: both matching conventions leave no-FFN accuracy at chance. Appendix~\ref{app:histogram-errors} shows that histogram strategies can shift internally, especially in the FFN-family, but the FFN remains necessary for the count readout or token-inventory component. Thus neither matching convention produces ablation-visible redistribution.
\end{enumerate}

\paragraph{Why we keep total-param matching as the headline.} Per-active matching is the standard convention in scaling work because its research question is whether sparsity can recover dense-level quality at lower active compute; there, the $4\times$ parameter excess is an accepted cost. Our research question inverts this: the per-token capacity reduction is the mechanism under study, and the relevant null hypothesis to rule out is ``MoE-GLU does better only because it has more total parameters.'' Reporting MoE-GLU at \emph{total-parameter} matching (same total params as dense, less per-active capacity) makes the redistribution claim conservative against this null: at parameter parity, MoE-GLU still redistributes substantially. The per-active-matched control above shows the effect persists in attenuated form even when we relax the parameter constraint to FLOP parity, ruling out the converse alternative that the effect requires the parameter excess.

\paragraph{Activation symmetry.} The per-active-matched MoE-GLU controls reported above use SiLU; activation-symmetric replicates at GELU (5 seeds each) yield no-FFN $16.8 \pm 4.2\%$ on add-7, $1.7 \pm 0.5\%$ on modadd, and $10.2 \pm 0.2\%$ on histogram, all within seed-to-seed standard deviation of the SiLU values (max $\Delta = 2$\,pp on add-7). The per-active matching robustness conclusion is therefore activation-invariant within measurement noise.

\FloatBarrier
\subsection{Auxiliary Loss Sweep}
\label{app:laux-sweep}

We sweep the load-balancing coefficient $\lambda_{\text{aux}} \in \{0, 10^{-3}, 10^{-2}, 10^{-1}, 1\}$ for MoE on modular addition.

\begin{figure}[htbp]
  \centering
  \includegraphics[width=0.85\linewidth]{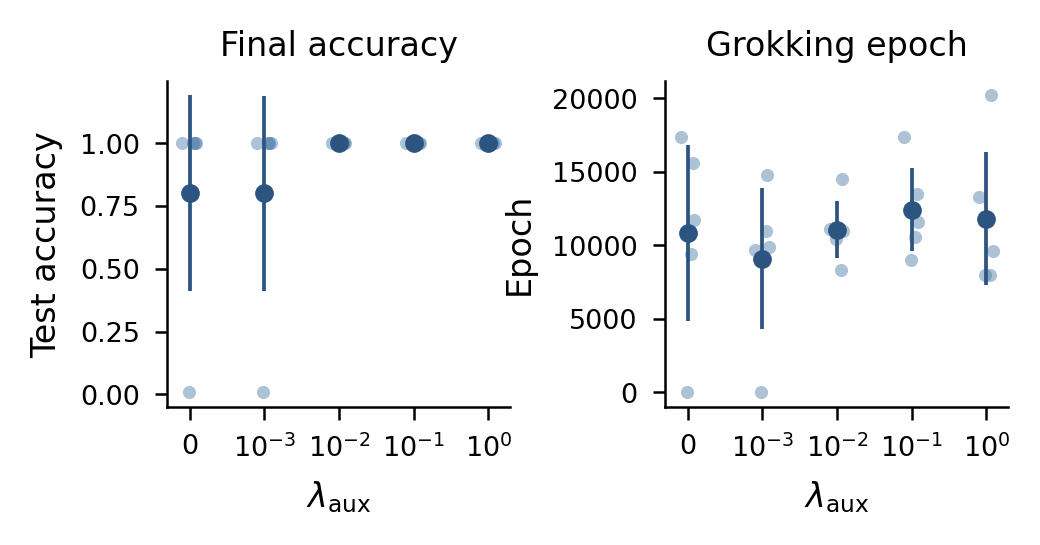}
  \caption{Auxiliary loss sweep for MoE on modular addition (5 seeds). \textbf{Left:}~Without load balancing ($\lambda_{\text{aux}} < 0.01$), 1/5 seeds fails to grok entirely. At $\lambda_{\text{aux}} \geq 0.01$, all seeds reach 100\%. \textbf{Right:}~Grokking epoch is comparable (${\sim}10$-$12$k) across all $\lambda_{\text{aux}}$ values; auxiliary loss determines \emph{whether} grokking occurs, not \emph{when}.}
  \label{fig:laux-sweep}
\end{figure}

%% ---- H: Histogram ----
\section{Histogram Error Modes and Controls}
\label{app:histogram-errors}

\paragraph{Random routing and narrow FFN on histogram.}
Consistent with histogram's limited ablation-visible redistribution, none of the controls produce a meaningful change in final accuracy (5 seeds each):

\begin{center}
\small
\begin{tabular}{lc}
\toprule
\textbf{Variant (histogram)} & \textbf{Test accuracy} \\
\midrule
Dense FFN (baseline)   & $99.96 \pm 0.02\%$ \\
Narrow FFN ($h{=}128$) & $99.51 \pm 0.56\%$ \\
MoE (learned routing)  & $99.73 \pm 0.15\%$ \\
MoE (random routing)   & $99.69 \pm 0.26\%$ \\
MoE-GLU (learned)      & $99.90 \pm 0.07\%$ \\
MoE-GLU (random)       & $99.92 \pm 0.02\%$ \\
\bottomrule
\end{tabular}
\end{center}

% When either component can solve the task alone (App.~\ref{app:frozen}: 96-99\% with frozen attention or frozen FFN), neither the routing choice nor the FFN width changes much.

These accuracy results do not imply that the internal strategy is unchanged. Rather, histogram requires complementary attention and FFN roles, so changes in internal strategy need not appear as improved robustness under component ablation.

\paragraph{Counting strategy diagnostics.}
To test whether these similar ablation outcomes hide different internal strategies, we probe where the count becomes linearly decodable. We train a linear classifier over count classes $\{0,\dots,L{-}1\}$ on the post-embedding stream, the post-attention residual, and the post-FFN residual. We also measure token-selective FFN structure by feeding token embeddings directly through the FFN's first projection and computing the fraction of neurons with strict token selectivity $s_i>5$.

\begin{table}[h]
\centering
\small
\begin{tabular}{lccccc}
\toprule
Variant & Post-attn acc & Attn lift & FFN lift & Sel{$>$}5 & Strategy \\
\midrule
FFN     & $0.755 \pm 0.081$ & $\mathbf{+0.572 \pm 0.083}$ & $+0.244 \pm 0.081$ & $0.33 \pm 0.03$ & Relation-style  \\
MoE     & $0.903 \pm 0.037$ & $\mathbf{+0.722 \pm 0.036}$ & $+0.092 \pm 0.037$ & $0.50 \pm 0.04$ & Relation-style  \\
GLU     & $0.282 \pm 0.017$ & $+0.101 \pm 0.022$ & $\mathbf{+0.717 \pm 0.017}$ & $0.93 \pm 0.02$ & Inventory-style \\
MoE-GLU & $0.366 \pm 0.053$ & $+0.178 \pm 0.049$ & $\mathbf{+0.629 \pm 0.053}$ & $0.96 \pm 0.01$ & Inventory-style \\
\bottomrule
\end{tabular}
\caption{Histogram counting strategy diagnostics over 5 seeds. \emph{Post-attn} is count-probe accuracy at the FFN input. \emph{Attn lift} is the gain over the embedding-only probe, and \emph{FFN lift} is the gain from the FFN module. \emph{Sel{$>$}5} is the fraction of FFN neurons exceeding a strict token-selectivity threshold. FFN-family variants expose more count information after attention, while GLU-family variants construct most count information through the FFN.}
\label{tab:histogram-strategy}
\end{table}

The diagnostic shows that histogram strategies differ by FFN architecture even when ablation outcomes are similar. In the FFN-family, MoE increases the relation-style attention contribution relative to dense FFN. In the GLU-family, the strategy remains inventory-style, with highly token-selective FFN structure preserved across experts. Thus histogram is not a case where the FFN is irrelevant. Instead, it is a case where attention and FFN roles remain complementary, so internal shifts do not translate into high no-FFN accuracy.

\paragraph{Matched-width GLU and MoE-GLU on histogram.}
For histogram, the primary GLU-family experiments already use the parameter-matched widths from Sec.~\ref{sec:experimental-setup}: GLU uses $h_{\text{glu}}{=}340$, giving $130{,}560$ GLU FFN parameters, which matches the dense FFN's $131{,}712$ parameters within $1\%$; MoE-GLU uses the corresponding expert width $h_E{=}85$. We report these results here to make the limited ablation-visible redistribution explicit under the same matched-width convention used in the main experiments (5 seeds each).

\begin{table}[h]
\centering
\small
\begin{tabular}{lccc}
\toprule
\textbf{Variant (histogram, $h{=}340$)} & Normal & No-attn & No-FFN \\
\midrule
GLU (param-matched)     & $100.0\%$ & $19.4 \pm 0.3\%$ & $9.8 \pm 0.1\%$ \\
MoE-GLU (param-matched) & $99.8\%$  & $18.6 \pm 0.5\%$ & $10.5 \pm 0.2\%$ \\
\bottomrule
\end{tabular}
\caption{Parameter-matched histogram controls (5 seeds each). Both variants show the same negligible no-FFN accuracy at parameter parity (gap $<1$\,pp).}
\label{tab:histogram-matched}
\end{table}

\noindent Histogram's low redistribution is task-imposed and survives architectural and parameter-matching manipulations, confirming it as a clean negative control for the add-7 and modadd results.

\begin{table}[htbp]
\centering
\small
\begin{tabular}{lccccccccc}
\toprule
& \multicolumn{9}{c}{\textbf{No-FFN Accuracy by Count Value}} \\
\cmidrule(lr){2-10}
\textbf{Variant} & 1 & 2 & 3 & 4 & 5 & 6 & 7 & 8 & 9 \\
\midrule
FFN     & 0\% & 4\% & 8\% & 47\% & 38\% & 14\% & 4\% & 0\% & 2\% \\
GLU     & 0\% & 2\% & 30\% & 62\% & 7\% & 0\% & 0\% & 0\% & 0\% \\
MoE     & 0\% & 0\% & 7\% & 27\% & 64\% & 25\% & 1\% & 0\% & 0\% \\
MoE-GLU & 0\% & 1\% & 40\% & 57\% & 3\% & 0\% & 0\% & 0\% & 0\% \\
\bottomrule
\end{tabular}
\caption{Histogram per-count accuracy under no-FFN ablation (5 seeds, all ${\sim}100\%$ normally). Extreme counts collapse across variants, while the surviving count band is architecture-dependent: FFN and MoE retain partial accuracy around counts 4--5, whereas GLU and MoE-GLU retain partial accuracy around counts 3--4. This pattern is consistent with softmax attention preserving some coarse count information while normalizing away exact count magnitude \citep{Ouellette2023CountingAA}.}
\label{tab:histogram-errors}
\end{table}

%% ---- I: Activation Function Robustness ----
\section{Activation Function Robustness}
\label{app:silu}

Many modern GLU-based architectures use SiLU rather than GELU gating. We train GLU and MoE-GLU with SiLU activation on all three tasks (5 seeds each) and rerun the core mechanistic analyses to verify our findings are not activation-specific.

\subsection{SiLU vs.\ GELU Comparison}
\label{app:silu-comparison}

\begin{figure}[htbp]
  \centering
  \includegraphics[width=0.85\linewidth]{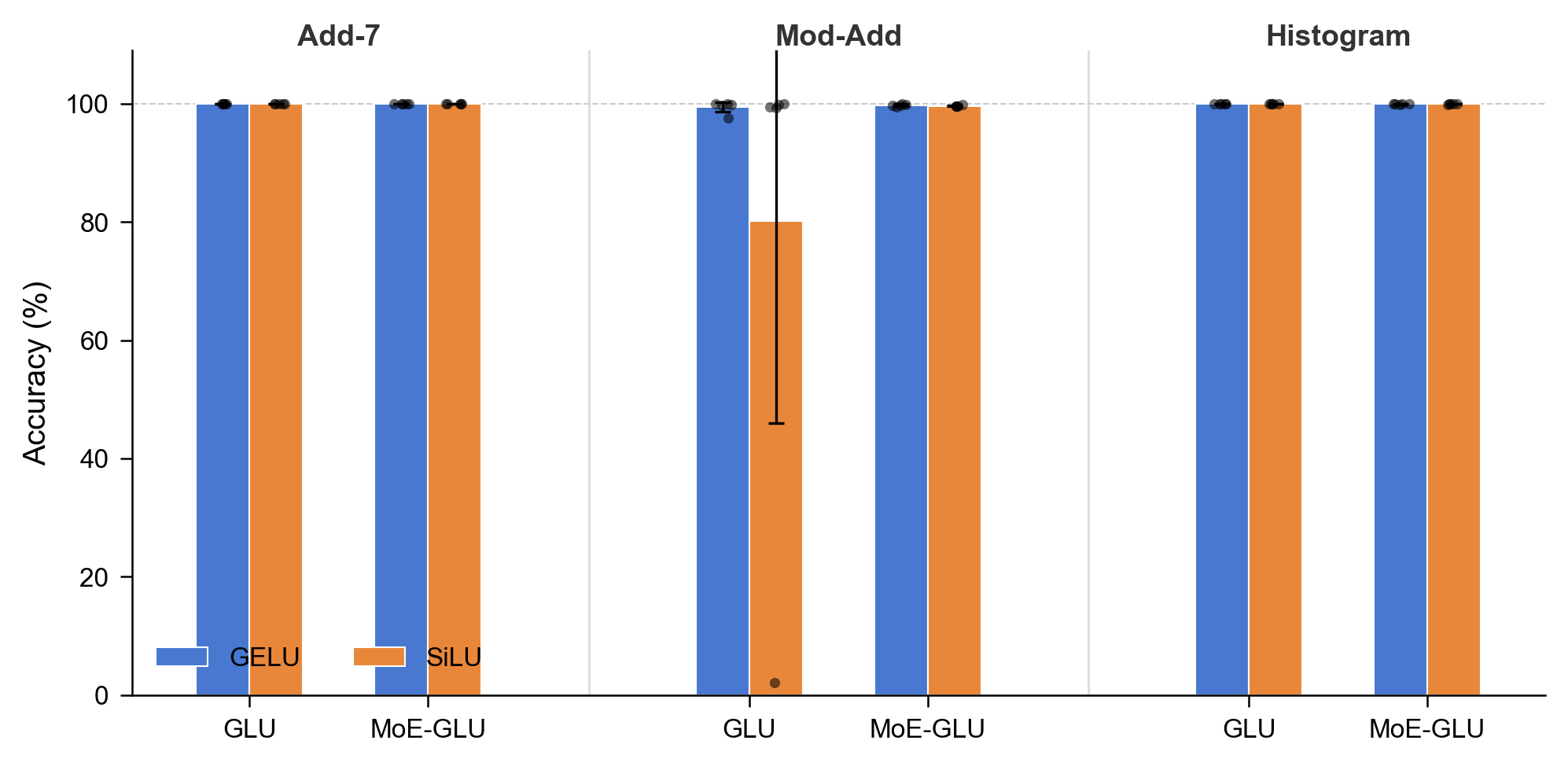}
  \caption{GELU vs.\ SiLU accuracy across all tasks and GLU variants (5 seeds, error bars: 95\% CI). SiLU matches GELU on add-7 and histogram. On modular addition, SiLU introduces a rare training instability: 1/5 GLU seeds fails to grok (final accuracy $31.5\%$ vs.\ ${\sim}99.6\%$ for the other four), pulling the 5-seed mean to $85.9\%$ in Tab.~\ref{tab:activation-symmetry}. MoE-GLU routing eliminates this instability (99.6\% vs.\ 99.7\% for GELU).}
  \label{fig:silu}
\end{figure}

\begin{figure}[htbp]
  \centering
  \includegraphics[width=0.85\linewidth]{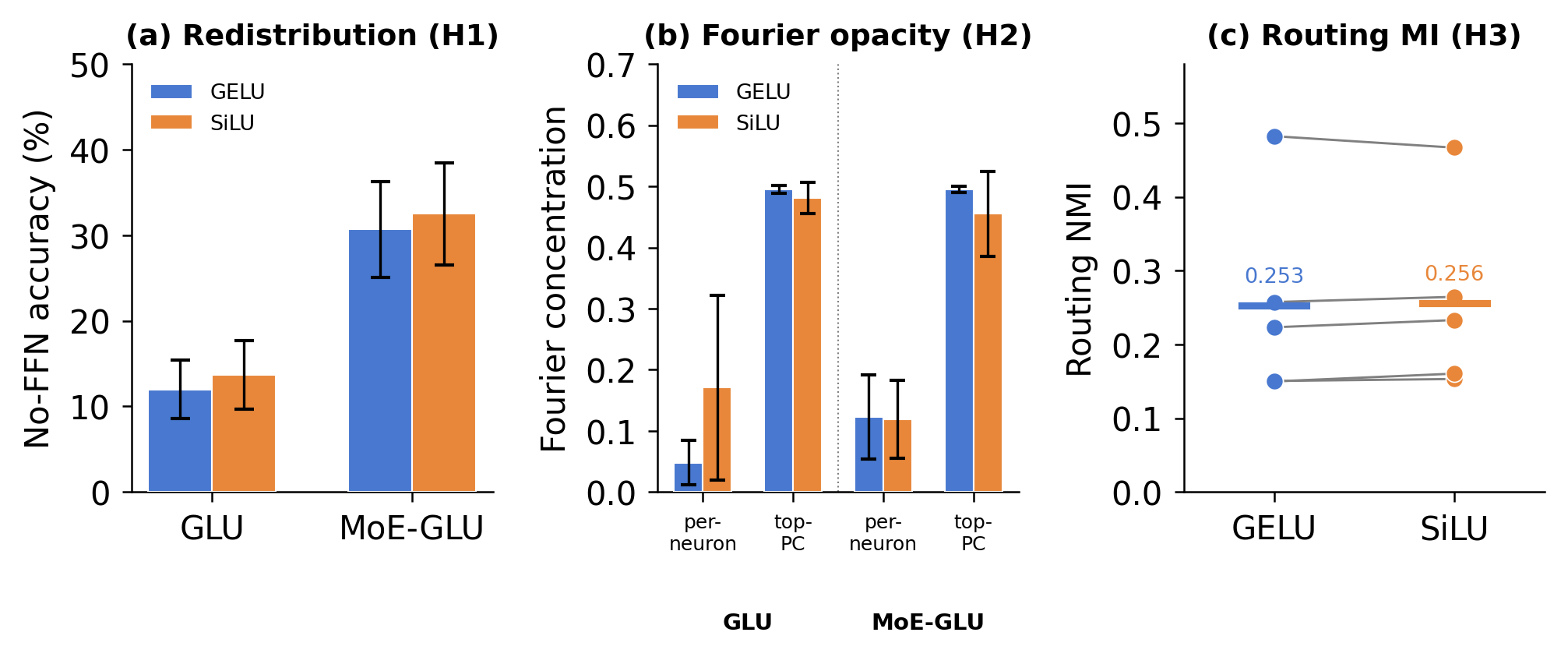}
  \caption{Mechanistic verification under SiLU. \textbf{(a)}~Redistribution: MoE-GLU's no-FFN survival advantage over GLU is preserved (${\sim}19$\,pp for both activations). \textbf{(b)}~Fourier opacity: per-neuron concentration stays low while top-PC concentration stays high for both activations, confirming the rotated representation. \textbf{(c)}~Routing MI: paired per-seed NMI values are virtually identical (means 0.253 vs.\ 0.256), confirming specialization is activation-agnostic.}
  \label{fig:silu-mechanisms}
\end{figure}

Beyond accuracy, we verify that all three hypotheses hold under SiLU (Fig.~\ref{fig:silu-mechanisms}). \textbf{H1 (redistribution):} MoE-GLU SiLU retains 32.5\% no-FFN accuracy vs.\ 13.7\% for GLU SiLU (+18.8\,pp), matching the GELU gap (+18.7\,pp). \textbf{H2 (Fourier opacity):} SiLU GLU per-neuron concentration is 0.17 with top-PC at 0.48 (GELU: 0.17 / 0.49, Tab.~\ref{tab:fourier-neuron-control}); the low-neuron, high-subspace pattern holds, though SiLU is noisier. \textbf{H3 (routing):} NMI between expert assignment and operation type is $0.256 \pm 0.114$ (SiLU) vs.\ $0.253 \pm 0.122$ (GELU), effectively identical, with per-seed correspondence confirming that specialization patterns are determined by initialization, not activation function.

\FloatBarrier
\subsection{Cross-Task Activation-Symmetric Ablations}
\label{app:activation-robustness}

To confirm that our findings are not artifacts of the gate nonlinearity, we ran every GLU and MoE-GLU experiment at both SiLU and GELU. The matching extends to FFN and plain MoE wherever a non-default activation was feasible. Across all paired (task, variant, activation) cells, the no-FFN ablation differs by less than the seed-to-seed standard deviation; the redistribution and Fourier findings are activation-invariant within measurement noise.

\begin{table}[h]
\centering
\small
\begin{tabular}{llcccc}
\toprule
\textbf{Task} & \textbf{Variant} & \textbf{Activation} & \textbf{Normal} & \textbf{No-attn} & \textbf{No-FFN} \\
\midrule
\textit{add-7 main} & FFN     & SiLU & 100.0\% & 32.3 $\pm$ 0.1 & 11.9 $\pm$ 3.6 \\
                    & FFN     & GELU & 100.0\% & 32.3 $\pm$ 0.0 & 10.8 $\pm$ 3.0 \\
                    & GLU     & SiLU & 100.0\% & 29.8 $\pm$ 5.0 & 13.7 $\pm$ 4.0 \\
                    & GLU     & GELU & 100.0\% & 29.8 $\pm$ 3.8 & 12.0 $\pm$ 3.4 \\
                    & MoE     & SiLU & 100.0\% & 16.8 $\pm$ 4.3 & 44.3 $\pm$ 12.5 \\
                    & MoE     & GELU &  99.9\% & 18.8 $\pm$ 5.1 & 43.8 $\pm$ 12.5 \\
                    & MoE-GLU & SiLU & 100.0\% & 21.8 $\pm$ 4.9 & 32.5 $\pm$ 6.0 \\
                    & MoE-GLU & GELU & 100.0\% & 25.7 $\pm$ 3.3 & 30.7 $\pm$ 5.6 \\
\midrule
\textit{add-7 matched ($h{=}170$)} & GLU    & SiLU & 100.0\% & 30.7 $\pm$ 3.0 & 9.7 $\pm$ 2.4 \\
                                   & GLU    & GELU & 99.9\%  & 30.7 $\pm$ 3.0 & 8.7 $\pm$ 1.5 \\
                                   & MoE-GLU& SiLU & 100.0\% & 21.8 $\pm$ 5.3 & 41.7 $\pm$ 2.8 \\
                                   & MoE-GLU& GELU & 99.9\%  & 20.3 $\pm$ 6.2 & 39.0 $\pm$ 3.8 \\
\midrule
\textit{hist main}  & FFN     & GELU & 99.9\% & 19.2 $\pm$ 1.2 & 11.5 $\pm$ 0.6 \\
                    & FFN     & SiLU & 99.9\% & 16.6 $\pm$ 1.6 & 11.5 $\pm$ 0.8 \\
                    & GLU     & GELU & 100.0\% & 19.7 $\pm$ 0.2 & 10.1 $\pm$ 0.3 \\
                    & GLU     & SiLU & 100.0\% & 19.7 $\pm$ 0.2 & 10.1 $\pm$ 0.3 \\
                    & MoE     & GELU & 99.7\% & 17.0 $\pm$ 1.4 & 12.0 $\pm$ 1.9 \\
                    & MoE     & SiLU & 99.5\% & 16.2 $\pm$ 1.3 & 11.7 $\pm$ 1.5 \\
                    & MoE-GLU & GELU & 99.9\% & 18.8 $\pm$ 0.5 & 10.0 $\pm$ 0.2 \\
                    & MoE-GLU & SiLU & 99.9\% & 18.7 $\pm$ 0.4 & 10.2 $\pm$ 0.4 \\
\midrule
\textit{hist matched ($h{=}340$)} & GLU     & GELU & 100.0\% & 19.4 $\pm$ 0.3 & 9.8 $\pm$ 0.1 \\
                                  & GLU     & SiLU & 100.0\% & 19.3 $\pm$ 0.5 & 9.9 $\pm$ 0.2 \\
                                  & MoE-GLU & GELU & 99.9\% & 18.6 $\pm$ 0.5 & 10.5 $\pm$ 0.2 \\
                                  & MoE-GLU & SiLU & 99.9\% & 18.3 $\pm$ 0.6 & 10.5 $\pm$ 0.2 \\
\midrule
\textit{modadd main} & FFN     & GELU & 66.8\% & 0.88 $\pm$ 0.00 & 1.34 $\pm$ 0.43 \\
                     & FFN     & SiLU & 79.7\% & 0.71 $\pm$ 0.35 & 1.30 $\pm$ 0.33 \\
                     & GLU     & GELU & 99.3\% & 0.88 $\pm$ 0.00 & 1.23 $\pm$ 0.30 \\
                     & GLU     & SiLU & 85.9\% & 0.88 $\pm$ 0.00 & 1.32 $\pm$ 0.39 \\
                     & MoE     & GELU & 99.6\% & 0.88 $\pm$ 0.00 & 7.48 $\pm$ 1.99 \\
                     & MoE     & SiLU & 99.6\% & 0.88 $\pm$ 0.00 & 6.32 $\pm$ 1.24 \\
                     & MoE-GLU & GELU & 99.6\% & 0.88 $\pm$ 0.00 & 3.71 $\pm$ 1.64 \\
                     & MoE-GLU & SiLU & 99.4\% & 0.88 $\pm$ 0.00 & 1.98 $\pm$ 0.83 \\
\midrule
\textit{modadd matched ($h{=}340$)} & GLU     & GELU & 99.4\% & 0.88 $\pm$ 0.00 & 1.40 $\pm$ 0.32 \\
                                    & GLU     & SiLU & 86.0\% & 0.88 $\pm$ 0.00 & 1.22 $\pm$ 0.18 \\
                                    & MoE-GLU & GELU & 90.8\% & 0.88 $\pm$ 0.00 & 2.57 $\pm$ 1.34 \\
                                    & MoE-GLU & SiLU & 99.6\% & 0.88 $\pm$ 0.00 & 2.78 $\pm$ 2.31 \\
\bottomrule
\end{tabular}
\caption{Activation-symmetric ablations (5 seeds each cell). Headline values reported in the main text correspond to the SiLU rows for add-7 and the GELU rows for modular addition and histogram, reflecting initial training defaults; the matched alternative activation is reported alongside in each block. For every (task, variant) pair with both activations measured, no-FFN accuracy differs by less than the seed-to-seed standard deviation. Maximum no-FFN gap under activation swap, taken across all (task, variant) pairs with both activations: FFN $1.1$\,pp, GLU $1.7$\,pp, MoE $1.16$\,pp, MoE-GLU $2.7$\,pp, all within per-seed std. The redistribution and Fourier findings are activation-invariant.}
\label{tab:activation-symmetry}
\end{table}

\noindent These numbers use the same digit-only metric (average over $o_0$ to $o_3$) as Fig.~\ref{fig:component-ablation} and Tab.~\ref{tab:ablation-summary}, so the add-7 rows here line up with the main-text values directly.

% \newpage
% \input{checklist.tex}

\end{document}